\documentclass{article}

\usepackage[main,final]{neurips_2026}
\usepackage[utf8]{inputenc}
\usepackage[T1]{fontenc}
\usepackage{hyperref}
\usepackage{url}
\usepackage{booktabs}
\usepackage{longtable}
\usepackage{array}
\usepackage{multirow}
\usepackage{enumitem}
\usepackage{dsfont}
\usepackage{amsfonts}
\usepackage{arydshln}
\usepackage{amsmath}
\usepackage{amssymb}
\usepackage{wrapfig}
\usepackage{nicefrac}
\usepackage{microtype}
\usepackage{graphicx}
\usepackage{algorithm, algpseudocode}
\usepackage[table]{xcolor}

\usepackage{tikz}
\usetikzlibrary{arrows.meta, positioning, calc, decorations.pathreplacing}
\usepackage{xcolor}
\usepackage[most]{tcolorbox}
\tcbuselibrary{skins,breakable}
\usepackage{fvextra}
\usepackage{pifont}
\usepackage{needspace}

\definecolor{tbPrompt}{HTML}{6C5BA7}
\definecolor{tbStudent}{HTML}{D9822B}
\definecolor{tbEnv}{HTML}{3B6EA8}
\definecolor{tbTeacher}{HTML}{2E8B57}
\definecolor{tbErr}{HTML}{C0392B}
\colorlet{tbErrHL}{tbErr!20}
\colorlet{tbFixHL}{tbTeacher!22}
\colorlet{tbReqHL}{yellow!45}

\newcommand{\tblegend}[2]{\mbox{\raisebox{-0.1ex}{\textcolor{#1}{\rule{0.9em}{0.9em}}}\,\,#2}}%
\newcommand{\tbcross}{\textcolor{tbErr}{\ding{56}}}%
\newcommand{\tbcheck}{\textcolor{tbTeacher}{\ding{52}}}%
\newcommand{\tbswitchicon}{\textcolor{white}{\ding{228}}}%
\newcommand{\tbarrow}{{\scriptsize$\rightarrow$}}
\newcommand{\tbcrossW}{\textcolor{white}{\ding{56}}}%
\newcommand{\tbcheckW}{\textcolor{white}{\ding{52}}}
\newcommand{\tbtagerr}{{\setlength{\fboxsep}{1.2pt}\colorbox{white}{\textcolor{tbErr}{\ding{56}\,STUDENT ERROR}}}}
\newcommand{\tbtagend}{{\setlength{\fboxsep}{1.2pt}\colorbox{white}{\textcolor{tbTeacher}{\ding{110}\,TASK COMPLETE, $T(\tau')=6$}}}}
\newcommand{\tbtagfix}{{\setlength{\fboxsep}{1.2pt}\colorbox{white}{\textcolor{tbTeacher}{\ding{52}\,TEACHER CORRECTION}}}}

\DefineVerbatimEnvironment{trajverb}{Verbatim}{%
  fontsize=\scriptsize, breaklines, breakanywhere,
  breaksymbolleft={}, breakindent=1.5em}
\newcommand{\tbMixColor}{tbErrHL}

\newtcolorbox{taskbox}[2][]{%
  enhanced, breakable,
  boxrule=0.7pt, arc=1.2mm, outer arc=1.2mm,
  left=2mm, right=2mm, top=3.2mm, bottom=1.2mm,
  before skip=3.5mm, after skip=2.5mm,
  fonttitle=\bfseries\footnotesize, coltitle=white,
  attach boxed title to top left={xshift=4mm, yshift*=-\tcboxedtitleheight/2},
  boxed title style={boxrule=0pt, arc=0.8mm, left=1.5mm, right=1.5mm, top=0.6mm, bottom=0.6mm},
  title={#2}, #1}

\tcbset{
  tbprompt/.style ={colframe=tbPrompt,  colback=tbPrompt!5,  boxed title style={colback=tbPrompt}},
  tbstudent/.style={colframe=tbStudent, colback=tbStudent!6, boxed title style={colback=tbStudent}, unbreakable},
  tbenv/.style    ={colframe=tbEnv,     colback=tbEnv!5,     boxed title style={colback=tbEnv}, unbreakable},
  tbteacher/.style={colframe=tbTeacher, colback=tbTeacher!6, boxed title style={colback=tbTeacher}, unbreakable},
  tberrcallout/.style={colframe=tbErr, colback=tbErr!7, borderline west={2.5pt}{0pt}{tbErr}},
  tbfixcallout/.style={colframe=tbTeacher, colback=tbTeacher!8, borderline west={2.5pt}{0pt}{tbTeacher}},
}

\newtcolorbox{tbswitch}{%
  enhanced, colback=black!85, colframe=black!85, coltext=white,
  arc=1.2mm, boxrule=0pt, left=3mm, right=3mm, top=1.6mm, bottom=1.6mm,
  before skip=4mm, after skip=3mm, halign=center, fontupper=\small}

\newtcolorbox{tbcallout}[1][]{%
  enhanced, boxrule=0.5pt, arc=1mm, left=2.5mm, right=2mm, top=1mm, bottom=1mm,
  before skip=1mm, after skip=3mm, fontupper=\footnotesize, #1}

\newcommand{\tbchip}[2]{{\setlength{\fboxsep}{1.6pt}\colorbox{#1}{\strut\textcolor{white}{\footnotesize #2}}}}
\newcommand{\tbswatch}[2]{{\setlength{\fboxsep}{1.2pt}\colorbox{#1}{\strut\footnotesize #2}}}
\newcommand{\tbswitchchip}[1]{{\setlength{\fboxsep}{1.6pt}\colorbox{black!85}{\strut\textcolor{white}{\footnotesize\ding{228}\,#1}}}\,}
\newtcolorbox{tbtimeline}{%
  enhanced, colback=black!3, colframe=black!25, boxrule=0.4pt, arc=1mm,
  left=1.5mm, right=1.5mm, top=1.2mm, bottom=1.2mm, halign=center}

\makeatletter
\newcommand{\tbfigcaption}[2]{\par\smallskip{\def\@captype{figure}\caption{#1}\label{#2}}\par}
\makeatother

\providecommand{\UseTaggingSocket}[1]{}
\providecommand{\SuspendTagging}[1]{}
\providecommand{\ResumeTagging}[1]{}

\usepackage[most]{tcolorbox}
\usepackage{listings}
\usepackage{caption}
\usepackage{hyperref}

\tcbuselibrary{skins, breakable, listings}

\definecolor{TaskColor}{HTML}{7B68A6}
\definecolor{EnvColor}{HTML}{3B86A8}
\definecolor{StudentColor}{HTML}{C97A2B}
\definecolor{TeacherColor}{HTML}{5A8F2B}

\lstdefinestyle{tracecode}{
  basicstyle=\ttfamily\scriptsize,
  breaklines=true,
  breakatwhitespace=false,
  columns=fullflexible,
  keepspaces=true,
  showstringspaces=false,
  numbers=none,
  frame=none,
  tabsize=2,
  upquote=true
}

\tcbset{
  trajectorybox/.style={
    enhanced,
    breakable,
    listing only,
    listing engine=listings,
    listing options={style=tracecode},
    boxrule=1.0pt,
    arc=0pt,
    outer arc=0pt,
    left=8pt,
    right=8pt,
    top=13pt,
    bottom=8pt,
    before skip=14pt,
    after skip=14pt,
    fonttitle=\bfseries\small,
    attach boxed title to top left={
      xshift=8mm,
      yshift*=-\tcboxedtitleheight/2
    },
    boxed title style={
      boxrule=1.0pt,
      arc=0pt,
      outer arc=0pt,
      left=6pt,
      right=6pt,
      top=2pt,
      bottom=2pt
    }
  }
}

\newtcblisting{TaskBox}[1][Task Description]{
  trajectorybox,
  title={#1},
  colback=TaskColor!9,
  colframe=TaskColor!85!black,
  boxed title style={
    colback=TaskColor!55,
    colframe=TaskColor!85!black
  }
}

\newtcblisting{EnvBox}[1][Environment]{
  trajectorybox,
  title={#1},
  colback=EnvColor!10,
  colframe=EnvColor!85!black,
  boxed title style={
    colback=EnvColor!45,
    colframe=EnvColor!85!black
  }
}

\newtcblisting{StudentBox}[1][Student Model Agent]{
  trajectorybox,
  title={#1},
  colback=StudentColor!10,
  colframe=StudentColor!85!black,
  boxed title style={
    colback=StudentColor!45,
    colframe=StudentColor!85!black
  }
}

\newtcblisting{TeacherBox}[1][Teacher Model Agent]{
  trajectorybox,
  title={#1},
  colback=TeacherColor!10,
  colframe=TeacherColor!85!black,
  boxed title style={
    colback=TeacherColor!45,
    colframe=TeacherColor!85!black
  }
}

\definecolor{delta}{HTML}{0F6E56}

\definecolor{grpoval}{HTML}{185FA5}

\definecolor{qGray}{HTML}{D9D9D9}
\definecolor{thinkLav}{HTML}{DDD0ED}
\definecolor{actionLav}{HTML}{C8B8E4}
\definecolor{envGreen}{HTML}{C8E6C9}
\definecolor{timeBlue}{HTML}{3070B0}
\newcommand{\tc}{\texttt{<tool\_call>}}
\newcommand{\tce}{\texttt{</tool\_call>}}
\newcommand{\tr}{\texttt{<tool\_response>}}
\newcommand{\tre}{\texttt{</tool\_response>}}
\newcommand{\ans}{\texttt{<answer>}}
\newcommand{\anse}{\texttt{</answer>}}

\newcommand{\Ci}{C_{\mathrm{i}}}
\newcommand{\Cg}{C_{\mathrm{tr}}}

\newcommand{\DG}{\mathcal{D}_{\mathcal{G}}}
\newcommand{\Dbc}{\mathcal{D}_{\mathrm{BC}}}
\newcommand{\Dop}{\mathcal{D}_{\mathrm{OP}}}

\newcommand{\pureBC}{\textsc{Pure-BC}}
\newcommand{\OPshort}{\textsc{OP-Short}}
\newcommand{\OPfull}{\textsc{OP-Full}}
\newcommand{\OPsuccess}{\textsc{OP-Success}}
\newcommand{\OPcritical}{\textsc{OP-Critical}}
\newcommand{\BC}{\textsc{BC}}

\newcommand{\Ki}{K_{\mathrm{i}}}
\newcommand{\Ktr}{K_{\mathrm{tr}}}
\newcommand{\PhiPre}{\Phi_{\mathrm{pre}}}
\newcommand{\PhiPost}{\Phi_{\mathrm{post}}}
\newcommand{\Tswitch}{\mathcal{T}_{\mathrm{switch}}}

\newcommand{\pie}{\pi_{e}}

\newcommand{\piOne}{\pi_{1}}
\newcommand{\Qtrain}{\mathcal{Q}_{\mathrm{train}}}
\newcommand{\Qtest}{\mathcal{Q}_{\mathrm{test}}}
\newcommand{\Tmax}{T_{\mathrm{max}}}
\newcommand{\ind}[1]{\mathds{1}\{#1\}}
\definecolor{cAxis1}{HTML}{7B2D8E}
\definecolor{cAxis2}{HTML}{2266AA}
\definecolor{cAxis3}{HTML}{228B22}

\definecolor{descgray}{gray}{0.50}

\colorlet{taxonomyHeadShade}{gray!15}
\colorlet{taxonomyRowShade}{gray!8}
\newcommand{\TaxonomyTableShading}{\rowcolors{2}{taxonomyRowShade}{white}}

\title{A Few Teacher Steps Go a Long Way: Efficient SFT Data Collection for LLM Agents}

\author{
  Junze Ye$^{1,}$\thanks{Equal contribution. Planned code release: https://github.com/junzeye/on-policy-sft} \quad
  Jiayi Cheng$^{2,*}$ \quad
  Miao Lu$^{1,*}$ \quad
  Michal Mankowski$^2$ \quad
  Jose Blanchet$^1$ \\
  \textbf{Mohsen Bayati}$^1$ \\[3pt]
  $^1$Stanford University \quad $^2$New York University \\[4pt]
  \texttt{\{junze,miaolu,jblanche,bayati\}@stanford.edu}, \texttt{\{jiayicheng,mankom01\}@nyu.edu}  
}

\begin{document}

\maketitle
\begin{abstract}
DAgger-style supervised finetuning (SFT) combines full teacher demonstrations with teacher continuations of partial student rollouts, providing supervision on contexts the student encounters. We study how to allocate the teacher-inference budget on these on-policy (OP) continuations after initial behavior cloning (BC). One existing collection recipe runs the teacher to termination and retains only successful continuations (OP-Success). Since rejected outputs still incur teacher-inference costs, we compare this recipe with shorter continuations without outcome filtering (OP-Short), and with continuing only failed learner rollouts while retaining successful continuations (OP-Critical). We evaluate these choices against acquiring additional full demonstrations, tracking total teacher-inference cost separately from retained training tokens. Across HotpotQA, ALFWorld, and OpenThoughts-TB-Dev, teacher continuations improve the accuracy-cost tradeoff over scaling full demonstrations. On HotpotQA and OpenThoughts-TB-Dev, OP-Short surpasses the strongest tested BC baseline while reducing estimated teacher-inference cost by 20\% and 83\%, respectively, and on ALFWorld five-step continuations cut teacher tokens by 22\% relative to full continuations with little accuracy loss. Controlled HotpotQA ablations show that by focusing on failed student rollouts, OP-Critical achieves comparable accuracy to OP-Success with 30\% lower estimated teacher-inference cost and 36\% fewer retained training tokens. Compute-matched comparisons with reinforcement learning and on-policy distillation further support brief teacher continuations as an efficient and lightweight post-training recipe.
\end{abstract}

\section{Introduction}
Supervised finetuning (SFT) is a simple and effective post-training method for large language models (LLMs), tuning a student model's weights to maximize the conditional likelihood of completions from human annotators \citep{ouyang2022} or a stronger teacher LLM \citep{alpaca}. For multi-turn LLM agents, standard SFT trains the student to imitate full start-to-end trajectories demonstrated by the teacher \citep{zeng2024agenttuning,openthoughts-agent}, an approach known in the imitation learning literature as \emph{behavior cloning} (BC) \citep{osa2018,finn2025cs224r-imitation}.

Compared to reinforcement learning from verifiable rewards (RL) \citep{tulu3} and on-policy distillation (OPD) \citep{agarwal2024gkd,gu2024minillm}, SFT provides dense token-level supervision without requiring teacher token probabilities, which are unavailable from human annotators or closed-weight API teachers. Still, collecting diverse teacher demonstrations at scale is expensive \citep{ouyang2022}, and when the pool of training tasks is fixed or costly to expand, performance from synthesizing more BC demonstrations often plateaus. This motivates our central question:
\begin{quote}
Given a pool of training tasks and environments, how should we spend additional teacher budget to expand an SFT dataset once behavior cloning plateaus?
\end{quote}
One natural baseline, which we denote \pureBC{}, is to collect more full start-to-end teacher demonstrations, optionally filtering for successful trajectories \citep{zelikman2022star,deepseekmath,acereason}. Because every context in \pureBC{} is induced by the teacher policy, however, adding more teacher rollouts never directly shows the student how to recover once its own mistakes push a trajectory into unfamiliar states \citep{ross2011dagger,caccia2024finetuningwebagents}.

Recent agentic post-training methods address this mismatch with DAgger-style data expansion \citep{ross2011dagger}: starting from an initial student trained on \pureBC{} data, they query the teacher to continue partial student rollouts and retrain on the combined dataset \citep{choudhury2024better,lyu2025score,lauffer2025oec,revisitDagger}. We refer to these teacher continuations as ``on-policy'' (\textsc{OP}) because they branch from contexts visited by the student policy. Within this family, existing recipes often run each teacher continuation to completion and keep only successful outcomes (\OPsuccess{}, motivated by \citealp{lauffer2025oec}). We introduce and study two more economical alternatives: \OPshort{}, which caps each teacher continuation at a few turns without outcome filtering, and \OPcritical{}, which queries the teacher only on failed student rollouts and retains successful continuations. Extended discussion of related work appears in Appendix~\ref{sec:literature}.

Comparing \textsc{OP} strategies' cost-performance frontier against \pureBC{}'s needs explicit cost accounting and matching. Prior DAgger-for-LLM studies often omit supervision cost or match only the number of retained training trajectories, which overlook differences in continuation length and the teacher inference spent on discarded rollouts. We instead track two complementary measures: $\Ci$, the total teacher inference spend, and $\Cg$, teacher-output tokens retained as SFT data. Separating $\Ci$ from $\Cg$ distinguishes the total expense of asking the teacher from the final training compute needs.

To study this design space systematically, we also describe \pureBC{} and \textsc{OP} recipes through data-collection hyperparameters covering switch timing, continuation length, filtering rules, and proposal counts. Varying these choices exposes a clear tradeoff across both state coverage and inference cost. Letting the teacher continue all the way to task completion spends budget on later states shaped mostly by the teacher, which the student is less likely to visit on its own. At the other extreme, stopping the teacher after a single turn spreads output tokens across many student prefixes, but may be too brief to show how to recover from an error, and forfeits KV-cache amortization: each takeover on a new student trajectory pays an uncached prefill over the full history so far, whereas continuing for a few consecutive turns reuses that cached prefix at much lower input cost.

We evaluate these tradeoffs on three agentic benchmarks: HotpotQA \citep{yang-etal-2018-hotpotqa} for search-based question answering, ALFWorld \citep{shridhar2021alfworld} for interactive text-world exploration, and OpenThoughts-TB-Dev (TB is short for Terminal-Bench) \citep{openthoughts-agent-original-blog} for command-line coding. In each benchmark, we first train an initial student on a seed \pureBC{} dataset of full teacher demonstrations. We use this initial student in two ways: to generate the rollouts from which \textsc{OP} prefixes are sampled, and as our baseline point for test accuracy and teacher cost prior to dataset expansion. Expanding from this baseline, \textsc{OP} continuations achieve substantially larger accuracy gains per unit of additional cost than collecting more full demonstrations (\pureBC{}). Within the \textsc{OP} family, sweeping continuation length favors short horizons: gains saturate around two to three teacher turns on HotpotQA, five turns outperform a single turn on ALFWorld, and three-turn continuations surpass scaled \pureBC{} at a fraction of the teacher cost on TB-Dev. Filtering sweeps on HotpotQA and ALFWorld show that this short-horizon advantage persists under outcome filtering: at three to five retained turns, \OPcritical{} matches unfiltered accuracy while cutting teacher-inference cost by 27\% to 42\% and retained training tokens by 47\% to 55\%.

Finally, we compare \OPshort{} with other online learning methods under matched compute resources. Starting from the same initial \pureBC{} student on HotpotQA, RL requires roughly 4 to $6\times$ the data-collection and SFT cost of an \OPshort{} checkpoint to reach its accuracy. On ALFWorld, where we host the teacher locally and can access token probabilities, reverse-KL OPD remains below \OPshort{} throughout a run using 9.9 active H100-hours, compared with 0.67 H100-hours for \OPshort{}. Together, these comparisons show that SFT on brief teacher continuations is an efficient option for offline-to-online learning, even when teacher token probabilities are available.

\paragraph{Paper organization.}
Section~\ref{sec:methods} formalizes our parameterized family of \pureBC{} and \textsc{OP} data-collection recipes alongside our dual cost metrics ($\Ci$ and $\Cg$). Section~\ref{sec:experiments} introduces our three research questions and experimental setups, and Section~\ref{sec:main-results} reports the corresponding accuracy-cost results across benchmarks. Section~\ref{sec:ablation} then compares \OPshort{} with reverse-KL OPD and RL under matched compute, before Section~\ref{sec:disc} discusses practical guidelines and limitations.
\section{Methodology}\label{sec:methods}
We consider ReAct-style agents \citep{react}, where an LLM policy interacts with an environment through an \emph{agent harness}: an orchestration wrapper that parses tool calls from each LLM step, executes them in the environment, and appends the formatted response to the LLM's context for the next agent step.
We model this agent loop as a finite-horizon partially observable Markov decision process (POMDP; \citealp{smallwood1973,kaelbling1998}), denoted by $\langle \mathcal{S},\mathcal{A},\mathcal{O},\mathrm{P}_s,\mathrm{P}_o,\mu,r\mid \Tmax,\mathcal{H}\rangle$, with environment states $\mathcal{S}$, agent steps $\mathcal{A}$, harness-formatted environment responses $\mathcal{O}$, LLM-readable contexts $\mathcal{H}$, and horizon $\Tmax$.

Here $\mu$ is a distribution over a task space $\mathcal{Q}$; in our experiments, $\mu_{\mathrm{train}}$ and $\mu_{\mathrm{test}}$ are empirical distributions over finite task sets $\Qtrain$ and $\Qtest$. Drawing a task $q\sim\mu$ initializes the episode with starting prompt context $h_0=(q)\in\mathcal{H}$ and initial environment state $s_0\in\mathcal{S}$ (and fixes the task's tools, transition kernels $\mathrm{P}_s,\mathrm{P}_o$, and reward $r$, whose dependence on $q$ we omit below). At each timestep $t=1,\ldots,\Tmax$,
\begin{equation}
a_t\sim\pi(\cdot\mid h_{t-1}),\qquad
s_t\sim\mathrm{P}_s(\cdot\mid s_{t-1},a_t),\qquad
o_t\sim\mathrm{P}_o(\cdot\mid s_t,a_t),
\label{eq:pomdp-dynamics}
\end{equation}
where $\pi\in\Pi$ is the LLM policy and each increment of $t$ executes one pass of Eq.~\ref{eq:pomdp-dynamics}. Here $a_t$ is a complete agent step (which may contain reasoning and multiple tool calls) and $o_t$ is the resulting environment response; the transition kernels $\mathrm{P}_s, \mathrm{P}_o$ may be deterministic.

In our experiments, the harness maintains an append-only \emph{context}:
\begin{equation}
h_t=h_{t-1}+(a_t,o_t)=(q,a_1,o_1,\ldots,a_t,o_t)\in\mathcal{H},
\label{eq:history}
\end{equation}
where $+$ denotes message concatenation and $h_0=(q)$ includes the initial system prompt, tool descriptions, and any initial observation. Unlike the environment state $s_t$, which may contain information unobservable to $\pi$, the context $h_t$ contains only the interaction history visible to the LLM. Appendix~\ref{app:rollout-anatomy} gives an illustrated example.

Every \emph{rollout} is a stochastic execution of the agent loop in Eq.~\ref{eq:pomdp-dynamics}. It terminates at a random stopping time $T\leq \Tmax$, realized either when the agent indicates task completion or when the turn budget $\Tmax$ is exhausted. At termination, the environment assigns a terminal reward $r(a_T,s_T)\in[0,1]$ from the final turn $a_T$ and state $s_T$; we do not consider process rewards in this paper.

A \emph{trajectory} $\tau$ is a realized path of the rollout process, with realized stopping time $T(\tau)$. In this work, we define
\begin{equation}
    \tau := \bigl((h_t,s_t,a_t,o_t)\bigr)_{t=0}^{T(\tau)},
    \label{eq:trajectory}
\end{equation}
where $a_0=o_0=\varnothing$ are initialization placeholders. During an ongoing rollout, only a trajectory prefix has been realized.
When ``trajectory'' is qualified by a variable name, it denotes the corresponding sequence, e.g., the \emph{action trajectory} is $(a_1,\ldots,a_{T(\tau)})$.

Along $\tau$, we write $h_t(\tau)$ and $s_t(\tau)$ for the LLM context and environment state at timestep $t$. We call their pair
$x_t(\tau):=(h_t(\tau),s_t(\tau))$ the \emph{rollout snapshot}
at time $t$, omitting $\tau$ when clear.

\subsection{Context Sampling and the SFT Objective}
\label{sec:context-sft-dist}
At the timestep level, an SFT data instance is a pair $(h,a^\star)$ of a context $h$ and a teacher-generated next agent turn $a^\star$. Unlike DAgger's fixed-horizon formulation \citep{ross2011dagger}, our context-sampling procedure must account for the random stopping time $T$.

For \textsc{OP}, we perform this sampling in two stages. First, draw a task $q\sim\mu$ and roll out a fixed student policy $\pi$ to obtain a completed trajectory $\tau$. Then, conditional on $\tau$, sample a \emph{switch time} $t'\sim\mathrm{Unif}\{1,\ldots,T(\tau)\}$ and take $h=h_{t'-1}(\tau)$ as the teacher's starting context. By contrast, for \pureBC{}, each full teacher trajectory $\tau^\star$ supplies a pair $(h_{t-1}(\tau^\star),a_t^\star)$ at every timestep $t \in \{1,\ldots,T(\tau^\star)\}$, and pooling all such pairs forms the empirical context distribution.

Different collection recipes generate training contexts in different ways, such as purely along teacher rollouts in \pureBC{} or from student prefixes (and any subsequent teacher turns) in \textsc{OP}, further reshaped by the filtering and truncation choices in \S\ref{sec:methods-data}. Each recipe therefore induces its own distribution over SFT contexts $h$, which we denote generically by $d$. Appendix~\ref{app:context-distribution} formalizes the sampling and weighting of initial teacher contexts, and Appendix~\ref{app:switch-time} contrasts our context sampling protocol with that of \citet{lauffer2025oec}. Given $d$ and a callable teacher $\pi_e$, we train a parameterized student $\pi_\theta$ by cross-entropy minimization,
\begin{equation}
\mathcal L_{\mathrm{SFT}}(\theta;d)
:=\mathbb E_{h\sim d}\!\left[
\mathbb E_{a^\star\sim\pi_e(\cdot\mid h)}
\left[-\log\pi_\theta(a^\star\mid h)\right]\right],
\label{eq:population-sft-ce}
\end{equation}
where the log-likelihood of $a^\star$ sums over its tokens. Minimizing Eq.~\ref{eq:population-sft-ce} over $\theta$ is equivalent to minimizing the expected \emph{forward} KL divergence,
\begin{equation}
\mathcal L_{\mathrm{FKL}}(\theta;d)
:=\mathbb E_{h\sim d}\!\left[
D_{\mathrm{KL}}\!\left(\pi_e(\cdot\mid h)\,\middle\|\,\pi_\theta(\cdot\mid h)\right)\right].
\label{eq:population-sft-kl}
\end{equation}
This differs from reverse-KL on-policy distillation (OPD), which minimizes $D_{\mathrm{KL}}(\pi_\theta\|\pi_e)$ \citep{zhao2026decoupling}. See Appendix~\ref{app:multiturn-sft} for the extension of the SFT loss formula to multi-turn teacher continuations.
\subsection{Data Collection Hyperparameters}
\label{sec:methods-data}
LLM frameworks such as \texttt{verl} \citep{verl} and \texttt{SGLang} \citep{zheng2024sglang} expose data-collection choices that we formalize as \emph{data hyperparameters}, determining both the empirical SFT context distribution and how supervision costs $(\Ci, \Cg)$ (\S\ref{sec:cost-model}) are spent. Given tasks $\Qtrain$, rollout student $\pi$, and teacher $\pie$, we collect data via
\begin{equation}
\mathcal G = (\Tswitch,\PhiPre,\PhiPost,\Ki,\Ktr,n).
\label{eq:dataset-recipe}
\end{equation}
For \textsc{OP} data collection, a proposal samples a student trajectory $\tau$ under $\pi$ and a switch time $t'\sim\Tswitch$ (e.g., as in \S\ref{sec:context-sft-dist}). If $t'>T(\tau)$ or the pre-continuation filter $\PhiPre$ rejects the prefix $\tau|_{t<t'}$, the proposal is aborted without querying $\pie$.
Otherwise, $\pie$ branches from snapshot $x_{t'-1}(\tau)$ and continues for $k_e \leq \Ki$ timesteps until task completion or the inference cap $\Ki$. This branch forms a hybrid trajectory $\tau'$ that shares $\tau$'s prefix through $t'-1$ ($x_{t'-1}(\tau')=x_{t'-1}(\tau)$) and evolves via teacher turns $a^\star_{t'},\ldots,a^\star_{t'+k_e-1}$ and the resulting environment observations $o'_{t'},\ldots,o'_{t'+k_e-1}$ along $\tau'$. A post-continuation filter $\PhiPost$ then screens $\tau'$; if accepted, its first $m=\min\{k_e,\Ktr\}$ teacher turns are retained for SFT. Truncating at $\Ktr<k_e$ is loosely analogous to clipping in PPO \citep{schulman2017ppo}, as each further teacher turn moves $\tau'$ away from the contexts the student visits.

This proposal process runs $n$ times per task in $\Qtrain$, with full pseudocode in Algorithm~\ref{alg:sft-rollout} (Appendix~\ref{app:data-collection-algorithm}). \pureBC{} is the special case with $t'=1$, no filtering, and $\Ki=\Ktr=\Tmax$, where $\pie$ starts from $h_0=q$ with no student involvement. Scaling \pureBC{} either keeps $\Qtrain$ fixed and increases $n$, or keep $n$ fixed and increases the size of $\Qtrain$, to add full teacher demonstrations.

Writing $\mathrm{SFT}(\pi_{\mathrm{init}}, \mathcal{D})$ for the policy obtained by initializing weights at $\pi_{\mathrm{init}}$ and minimizing empirical cross-entropy (Eq.~\ref{eq:population-sft-ce}) over dataset $\mathcal{D}$, every \textsc{OP} recipe $\mathcal G$ first trains an initial student $\pi_1$ from the base model $\pi_0$ on a seed \pureBC{} dataset $\Dbc$:
\begin{equation}
    \pi_1 := \mathrm{SFT}(\pi_0,\Dbc).
\label{eq:pi-1}
\end{equation}
Holding $\pi=\pi_1$ fixed, we initialize a buffer $\DG\gets\Dbc$ (or a task-matched subset of $\Dbc$ on TB-Dev; see \S\ref{sec:lab-data}) and aggregate each accepted \textsc{OP} proposal into $\DG$:
\begin{equation}
\DG \gets \DG \cup \Bigl\{\bigl(
\underbrace{h_{t'-1}(\tau)}_{\text{student context}},\;
\underbrace{(a^\star_{t'},o'_{t'},\ldots,a^\star_{t'+m-1},o'_{t'+m-1})}_{\text{retained teacher continuation from }\tau'}
\bigr)\Bigr\}.
\label{eq:D_g}
\end{equation}
When training on $\DG$, cross-entropy loss is computed only on the teacher turns $a^\star_{t'},\ldots,a^\star_{t'+m-1}$, while $h_{t'-1}(\tau)$ and $o'_{\cdot}$ appear only in the conditioning context without loss. Rejected proposals are not replaced, so $\Delta|\DG|\le n|\Qtrain|$. After one collection round ($\DG=\Dbc\cup\Dop$, with $\Dop$ denoting the newly collected continuations), we train the final model $\pi_2$ reinitialized from $\pi_0$:
\begin{equation}
    \pi_2 := \mathrm{SFT}(\pi_0,\DG).
\label{eq:pi-2}   
\end{equation}
Table~\ref{tab:design-space} summarizes the \pureBC{} and \textsc{OP} recipes we study alongside prior methods. We compare recipes $\mathcal G$ by evaluating $\pi_2$ on $\Qtest$; from \S\ref{sec:experiments} onward we refer to $\pi_2$ by its recipe name (e.g., \OPshort{}, or generically an \textsc{OP} model), keeping in mind that $\DG$ combines seed \pureBC{} data with $\Dop$.

\begin{table}[H]
  \centering

  \caption{Hyperparameter recipes under Eq.~\ref{eq:dataset-recipe}.
  ``All'' means no filtering. The OP rows use $\pi=\pi_1$. Setting
  $\Ktr=\Ki$ retains every teacher turn in an accepted continuation.
  OEC \citep{lauffer2025oec} and LEAP/SCoRe
  \citep{choudhury2024better,lyu2025score} are included for reference.}
  \label{tab:design-space}

  \small
  \setlength{\tabcolsep}{5.3pt}
  \renewcommand{\arraystretch}{1.15}
  \vspace{4pt}
  \TaxonomyTableShading
  \begin{tabular}{lcccccc}
  \toprule
  \rowcolor{taxonomyHeadShade}
  \textbf{Recipe}
    & $\Tswitch$ & $\PhiPre$ & $\PhiPost$
    & $\Ki$ & $\Ktr$ & $n$ \\
  \midrule

  \pureBC
    & $t'=1$ & All & All
    & $\Tmax$ & $=\Ki$ & $1,2,\ldots$ \\
  \addlinespace[2pt]
  \OPshort
    & Random & All & All
    & swept & $=\Ki$ & fixed \\
  \addlinespace[2pt]
  \OPfull
    & Random & All & All
    & $\Tmax$ & $=\Ki$ & fixed \\
  \addlinespace[2pt]
  \OPsuccess
    & Random & All & Teacher success
    & $\Tmax$ & swept$^{\dagger}$ & fixed \\
  \addlinespace[2pt]
  \OPcritical
    & Random & Student failure & Teacher success
    & $\Tmax$ & swept$^{\dagger}$ & fixed \\
  \midrule

  OEC
    & $\mathrm{Unif}\{1,\ldots,31\}$
    & All
    & \shortstack{Teacher success\\\& no action repetition}
    & $\Tmax$ & $=\Ki$ & fixed \\
  \addlinespace[2pt]
  LEAP/SCoRe
    & LLM-teacher-judged & Student failure & All
    & $1$ & $1$ & fixed \\
  \bottomrule
  \end{tabular}

  \vspace{2pt}
  \begin{minipage}{\textwidth}
  \footnotesize
  ``Random'' in our OP rows means
  $t'\sim\mathrm{Unif}\{1,\ldots,T(\tau)\}$.
  A rule-based verifier determines success or failure from the terminal
  outcome of $\tau$ (student) or $\tau'$ (teacher).
    OEC samples from a fixed horizon of 31 and discards proposals with $t' > T(\tau)$, motivating our invalid-switch rule in \S\ref{sec:methods-data}.

  \par\smallskip
  $^{\dagger}$ After outcome filtering, we vary retained teacher length
  $\Ktr\in\{1,2,3,\Ki\}$ on HotpotQA and
  $\Ktr\in\{1,3,5,\Ki\}$ on ALFWorld (\S\ref{sec:lab-data}).
  Terminal-Bench-Dev evaluates unfiltered recipes only.
  \end{minipage}
\end{table}

\subsection{Cost Accounting for SFT Teacher Supervision}
\label{sec:cost-model}
For each recipe $\mathcal G$ in \S\ref{sec:methods-data}, our primary cost measure is \emph{total teacher inference cost} $(\Ci)$, including generations rejected by $\PhiPost$. We also report the number of teacher-generated tokens retained for SFT $(\Cg)$, a proxy for supervision volume per training epoch. Both include the seed data $\Dbc$. While OEC \citep{lauffer2025oec} and SCoRe \citep{lyu2025score} report token usage statistics, their comparisons between BC and OP do not evaluate performance against total teacher-inference expense.

With the proposal budget $n$ fixed, rejection by $\PhiPre$ avoids querying the teacher, whereas $\PhiPost$ discards supervision after its inference cost has been incurred. The teacher inference step cap $\Ki$ affects both cost measures, whereas lowering $\Ktr$ alone reduces $\Cg$ while leaving $\Ci$ unchanged.

In our experiments, $\Ci$ is estimated as API expense in US dollars for teachers queried via OpenRouter, or as total teacher output tokens for locally hosted models. For API-based teachers, we reconstruct each call's input context and output from locally saved conversation logs and compute
\begin{equation}
\Ci =
p_{\mathrm{in}}(N_{\mathrm{in}} - N_{\mathrm{cache}})
+ p_{\mathrm{cache}}N_{\mathrm{cache}}
+ p_{\mathrm{out}}N_{\mathrm{out}}.
\end{equation}
Here $N_{\mathrm{in}}$ and $N_{\mathrm{out}}$ count input and output tokens across all teacher calls before filtering, $N_{\mathrm{cache}}$ is the estimated count of KV-cached input tokens, and $p_{\mathrm{in}}, p_{\mathrm{cache}}, p_{\mathrm{out}}$ denote the corresponding public API prices in USD per token (counted with the teacher's chat template and tokenizer). Among major API providers, $p_{\mathrm{out}}$ is typically 4 to $5\times$ $p_{\mathrm{in}}$, while $p_{\mathrm{cache}}$ is around $0.1\times$ $p_{\mathrm{in}}$.

By Eq.~\ref{eq:history}, consecutive teacher API calls share the preceding call's input context as an exact prefix. We estimate $N_{\mathrm{cache}}$ from these shared prefixes, weighted by a cache-hit rate, with no cache credit on the first call of each teacher rollout or continuation. Consequently, the switch-time distribution $\Tswitch$ directly affects inference cost: an \textsc{OP} continuation may begin from a long student-generated context and incur a large initial prefill charge, whereas a \pureBC{} teacher begins from the shorter task prompt. Allowing the teacher to continue multiple agent steps ($\Ki > 1$) helps amortize this initial prefill across subsequent cached calls. Whether \textsc{OP}'s learning gains outweigh the net prefill overhead is an empirical question we investigate.

\section{Experimental Setup}\label{sec:experiments}
Our main experiments study three questions on the tradeoff between accuracy and supervision cost:
\begin{enumerate}[label=\textbf{Q\arabic*.}, leftmargin=*]
    \item Are teacher continuations from student-induced contexts (\textsc{OP}) more effective than additional full teacher demonstrations (\pureBC{}) once we account for total teacher inference cost $\Ci$?
    \item When branching from a student-induced context, should the teacher continue for a single step, a few steps, or until task completion ($\Ki$)?
    \item How do trajectory filtering ($\PhiPre, \PhiPost$) and truncating retained continuations ($\Ktr$) affect efficiency with respect to $\Ci$ and $\Cg$?
\end{enumerate}

We evaluate these questions across three agent environments (\S\ref{sec:lab-env}), using the student-teacher pairs and data collection protocols detailed in \S\ref{sec:lab-data}.

\subsection{Environments}
\label{sec:lab-env}
\textbf{HotpotQA} \citep{yang-etal-2018-hotpotqa} is a multi-hop question answering benchmark in which an LLM synthesizes evidence across multiple Wikipedia passages to produce a verifiable answer. We build on the search harness from Search-R1 \citep{searchr1}, implemented in \texttt{verl} \citep{verl}, which lets the agent iteratively query an archived Wikipedia corpus. Whenever a valid search query is parsed from an agent step, the environment returns three retrieved Wikipedia excerpts; malformed queries return an error prompt asking the agent to retry. Training tasks are drawn from the \texttt{train-medium} split, and test accuracy is evaluated on the official $7{,}405$-task \texttt{fullwiki} split used by Search-R1.

\textbf{OpenThoughts-TB-Dev} \citep{openthoughts-agent-original-blog}, a.k.a. TB-Dev, is a benchmark of $70$ command-line tasks curated by the OpenThoughts-Agent project to track Terminal-Bench 2.0 \citep{merrill2026terminal}. We reserve TB-Dev strictly for final test evaluation. Training tasks instead come from \href{https://huggingface.co/datasets/open-thoughts/OpenThoughts-Agent-v1-SFT}{\texttt{OpenThoughts-Agent-v1-SFT}} \citep{openthoughts-agent}, denoted \texttt{v1-SFT}, which provides $15{,}209$ coding tasks from \texttt{nl2bash} \citep{nl2bash} and \texttt{inferredbugs} \citep{inferifx2023} with one \texttt{GLM-4.6} teacher demonstration per task. Both teacher and student policies are rolled out with the \texttt{Terminus-2} agent harness \citep{merrill2026terminal} inside task-specific Docker CPU sandboxes.

\textbf{ALFWorld} \citep{shridhar2021alfworld} is a suite of text-based household simulators. At the start of an episode, the agent receives only a brief text description of the room layout and its goal (e.g., ``take a sandwich from the fridge and heat it up on the stove''). To succeed, it must explore rooms and receptacles to locate target objects before acting on them. We integrate ALFWorld into our \texttt{verl} fork with a custom agent loop, adapting prompt templates and formatting from \texttt{verl-agent} \citep{verl-agent}. Test accuracy is evaluated on the $134$-task \texttt{valid-unseen} split, which consists of household layouts unseen during training.

\subsection{Data Collection Settings}
\label{sec:lab-data}
\begin{table}[htbp]
  \centering
  \caption{LLM policies and cost settings across environments.}
  \label{tab:data-settings-1}

  \small
  \setlength{\tabcolsep}{3pt}
  \renewcommand{\arraystretch}{1.15}
  \def\BCSettingsWidth{\dimexpr\linewidth-10\tabcolsep\relax}
  \vspace{4pt}
  \TaxonomyTableShading
  \begin{tabular}{@{}
    >{\raggedright\arraybackslash}p{0.20\BCSettingsWidth} %
    >{\centering\arraybackslash}p{0.07\BCSettingsWidth}   %
    >{\raggedright\arraybackslash}p{0.19\BCSettingsWidth} %
    >{\raggedright\arraybackslash}p{0.26\BCSettingsWidth} %
    >{\centering\arraybackslash}p{0.17\BCSettingsWidth}   %
    >{\centering\arraybackslash}p{0.11\BCSettingsWidth}   %
    @{}}
    \toprule
    \rowcolor{taxonomyHeadShade}
    \textbf{Environment}
      & \textbf{$\Tmax$}
      & \textbf{Student $\pi_0$}
      & \textbf{Teacher $\pie$}
      & \textbf{Teacher access}
      & \textbf{$\Ci$ unit} \\
    \midrule
    HotpotQA
      & $5$
      & \texttt{Qwen2.5-\allowbreak3B}
      & \texttt{Qwen3-235B-\allowbreak A22B}
      & API
      & USD \\
    \addlinespace[2pt]
    TB-Dev (\texttt{v1-SFT})
      & $32$
      & \texttt{Qwen3-8B}
      & \texttt{GLM-4.6}
      & API
      & USD \\
    \addlinespace[2pt]
    ALFWorld
      & $50$
      & \texttt{Qwen2.5-\allowbreak0.5B}
      & \texttt{Qwen3-8B} (after RL)
      & Local inference
      & tokens \\
    \bottomrule
  \end{tabular}
\end{table}

Table~\ref{tab:data-settings-1} lists each environment's rollout horizon $\Tmax$ and model configuration, including how $\Ci$ is measured. As in \S\ref{sec:methods-data}, every \textsc{OP} run generates student prefixes with $\pi_1 = \mathrm{SFT}(\pi_0, \Dbc)$ and samples one uniform switch time per rollout. Per Table~\ref{tab:design-space}, \OPfull{} sets $\Ki=\Ktr=\Tmax$, whereas \OPsuccess{} and \OPcritical{} run full teacher continuations ($\Ki=\Tmax$) before outcome filtering and truncate retained continuations to $\Ktr$ teacher steps. After data collection and SFT training, we run test evaluations of each trained checkpoint with multiple random seeds and take their average (8, 8, and 3 trials per task for HotpotQA, ALFWorld, and TB-Dev).

\textbf{HotpotQA.} We randomly sample $1{,}000$ tasks from the \texttt{train-medium} split and fix them as $\Qtrain$ across all runs. Then, we generate four i.i.d. teacher demonstrations per $\Qtrain$ task and form nested \pureBC{} baselines from the first $n\in\{1,2,3,4\}$ demonstrations. The $n=1$ set forms $\Dbc$ and is included in every \textsc{OP} training set. In \textsc{OP}, we collect $n_{\textsc{OP}}=3$ rollouts per task with $\pi_1$ ($3{,}000$ proposals before filtering), comparing \OPshort{} at $\Ki\in\{1,2,3\}$ against \OPfull{} ($\Ki=\Tmax=5$), and sweeping $\Ktr\in\{1,2,3,\Tmax\}$ for \OPsuccess{} and \OPcritical{}.

\textbf{TB-Dev.} We shuffle the $15{,}209$ demonstrations in \texttt{v1-SFT} once and form nested \pureBC{} datasets from the first $\{10,25,50,75,100\}\%$ ($1{,}520$ to $15{,}209$ demonstrations). Because \texttt{v1-SFT} has one demonstration per task, scaling \pureBC{} here expands task coverage. The $10\%$ subset ($1{,}520$ demonstrations) forms $\Dbc$ and trains $\pi_1$. On $760$ tasks\footnote{Earlier versions of this paper had a typo of $720$ tasks. The reported costs and evaluation scores were nonetheless based on the full $760$ tasks, so this correction does not change the direction of our conclusions.} from this subset, we collect $n=1$ unfiltered \textsc{OP} continuation per task under \OPshort{} ($\Ki \in \{1,3\}$) or \OPfull{} ($\Ki=\Tmax=32$). Final SFT (Eq.~\ref{eq:pi-2}) uses the matching $760$ seed demonstrations paired with the corresponding \textsc{OP} continuations, yielding $1{,}520$ training examples; teacher-inference cost $\Ci$ still counts all $1{,}520$ demonstrations used to train $\pi_1$.

\textbf{ALFWorld.} We fix $\Qtrain$ at $32$ tasks, sampled from the \texttt{train} split and balanced across the six task types. The teacher policy $\pie$ is a \texttt{Qwen3-8B} model that we previously RL-finetuned on tasks in the full \texttt{train} split, and on the 32-task $\Qtrain$ it reaches a success rate of $92.7\%$. We form nested \pureBC{} datasets with $n\in\{32,48,64\}$ demonstrations per task; the $32\times32$ set ($1{,}024$ demonstrations) forms $\Dbc$ and is included in every \textsc{OP} training set. We collect $n_{\textsc{OP}}=64$ rollouts per task from $\pi_1$ on $\Qtrain$, then compare \OPshort{} at $\Ki\in\{1,3,5\}$ against \OPfull{} ($\Ki=\Tmax=50$), and sweep $\Ktr\in\{1,3,5,\Tmax\}$ for \OPsuccess{} and \OPcritical{}.

Further details about the data collection and inference setup, SFT hyperparameters and checkpoint-selection protocol are provided in Appendix~\ref{app:exp-details}.

\section{Main Results}\label{sec:main-results}
We organize our main empirical findings around the three questions posed in \S\ref{sec:experiments}. We first compare unfiltered \textsc{OP} continuations against scaling \pureBC{} under total teacher-inference cost $\Ci$ (Q1, \S\ref{sec:q1-OP-BC-efficiency}), and then vary the continuation horizon $\Ki$ from a single teacher step to full rollouts (Q2, \S\ref{sec:q2-teacher-horizon}). We conclude by evaluating how rollout filtering ($\PhiPre, \PhiPost$) and continuation truncation ($\Ktr$) shape both $\Ci$ and retained training tokens $\Cg$ (Q3, \S\ref{sec:q3-filtering-truncation}).
Total teacher-inference cost $\Ci$ includes the seed-BC set $\Dbc$ and is measured as (i) estimated API expense in US dollars under $0.1\times$ cache-read pricing for HotpotQA and TB-Dev, or (ii) total teacher output tokens for ALFWorld.

\subsection{Is OP more efficient than \pureBC{} in terms of teacher cost?}\label{sec:q1-OP-BC-efficiency}

\begin{figure}[!htbp]
\centering
\captionsetup{font=normalsize,labelfont=bf,skip=3pt,singlelinecheck=false}
\includegraphics[width=\linewidth]{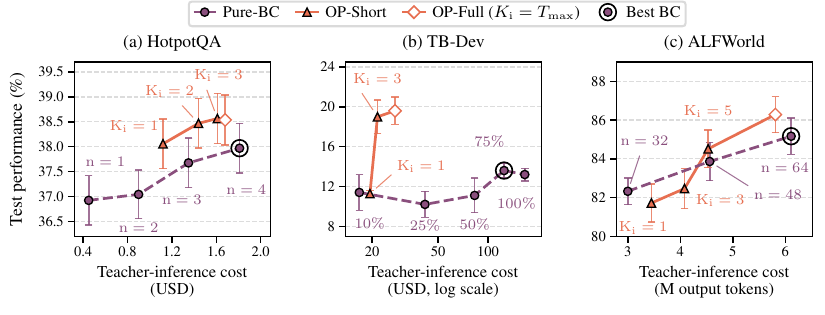}
\caption{Teacher-inference cost and test performance.
\pureBC{} scales the demonstration budget, while \OPshort{} varies $\Ki$.
Hollow diamonds show \OPfull{}.
Error bars show one standard error.
OP costs include seed-BC collection.
The leftmost \pureBC{} point in each subfigure is $\pi_1$.}
\label{fig:q1-cost-performance}
\par\vspace{8pt}

\begingroup
  \captionsetup{font=normalsize,labelfont=bf,skip=3pt,hypcap=false,singlelinecheck=false}
  \captionof{table}{Mean OP-over-BC accuracy gains [paired 95\% CIs], in pp, and teacher-cost reductions.
  ``Best BC'' is the \pureBC{} checkpoint with the highest test accuracy among the baselines.}
  \label{tab:q1-summary}
  \small
  \setlength{\tabcolsep}{5pt}
  \renewcommand{\arraystretch}{1.15}
  \noindent
  \begin{tabular*}{\linewidth}{@{\extracolsep{\fill}}lcccc@{}}
    \toprule
    Benchmark & $K_{\mathrm{i}}$ & vs.\ initial BC ($\pi_1$) & vs.\ best BC & Cost reduction \\
    \midrule
    HotpotQA & $2$        & $\mathbf{+1.55}\;[1.19,\,1.91]$  & $\mathbf{+0.50}\;[0.14,\,0.86]$  & $20.4\%$ \\
    \addlinespace[2pt]
    TB-Dev   & $3$        & $\mathbf{+7.65}\;[1.77,\,13.65]$ & $\mathbf{+5.43}\;[0.43,\,10.41]$ & $82.7\%$ \\
    \addlinespace[2pt]
    ALFWorld & $T_{\max}$ & $\mathbf{+3.96}\;[1.45,\,6.53]$  & $\mathbf{+1.12}\;[-1.49,\,3.82]$ & $4.9\%$ \\
    \bottomrule
  \end{tabular*}
  \par
\endgroup

\end{figure}

Figure~\ref{fig:q1-cost-performance} compares the test accuracy of checkpoints trained on \pureBC{}
versus \OPshort{} or \OPfull{} datasets, without extra student context or teacher continuation filtering ($\PhiPre = \varnothing = \PhiPost$).
This setup aims to attribute performance changes to mainly the choice of $\Tswitch$.
The \pureBC{} curves scale the demonstration budget $n$, while the OP curves vary $\Ki$.
To identify the $\Ki$ threshold in Table~\ref{tab:q1-summary}, we seek a clear accuracy gain over
the initial BC policy $\pi_1$ and accuracy on par or better than the strongest \pureBC{} baseline.
We assess each gain with a paired bootstrap CI over test tasks: we average each
model's repeated evaluations within a task, then resample tasks with replacement two million times, keeping both models' scores for a sampled task together, and
recompute the difference in mean accuracy.
A gain is clear when the resulting 95\% CI lies above zero.

With sufficiently long teacher continuations, adding $\Dop$ meets both goals
in all three environments. On HotpotQA and TB-dev, $\Ki=2$ and $3$ are the
smallest tested caps with clear gains over both BC baselines. On ALFWorld,
full continuations first give a clear gain over $\pi_1$ and attain comparable
accuracy to best BC. All three selected OP models have higher mean accuracy
than both references, with teacher-cost reductions of $20.4\%$, $82.7\%$, and
$4.9\%$, respectively, without any filtering.

\subsection{How many teacher steps are needed?}
\label{sec:q2-teacher-horizon}

To answer Q2, we vary the teacher inference step cap $\Ki$ while keeping the student policy $\pi_1$ and the proposal budget $n$ fixed.
Table~\ref{tab:q2-teacher-horizon} compares stopping teacher continuations after one step or a few steps with letting the teacher continue to completion, without outcome filtering.
Figure~\ref{fig:q1-cost-performance} plots the continuation-length sweep alongside the \pureBC{} scaling curve.

\begin{table}[htbp]
  \centering
  \caption{\emph{Teacher continuation length and the cost-performance tradeoff.}
  Mean accuracy differences [95\% CIs], in pp.
  Each cost reduction ($\Ci$) is a percentage of the \OPfull{} cost, and each cost increase ($\Ci$) is a percentage of the \OPshort{} ($\Ki=1$) cost.}
  \label{tab:q2-teacher-horizon}
  \small
  \setlength{\tabcolsep}{3.5pt}
  \renewcommand{\arraystretch}{1.12}
  \begin{tabular}{@{}lccrcr@{}}
    \toprule
    & & \multicolumn{2}{c}{Versus \OPfull{}} & \multicolumn{2}{c}{Versus \OPshort{} ($\Ki=1$)} \\
    \cmidrule(lr){3-4}\cmidrule(l){5-6}
    Benchmark & $\Ki$ & Difference [95\% CI] & $\Ci$ reduction & Difference [95\% CI] & $\Ci$ increase \\
    \midrule
    \multirow{2}{*}{HotpotQA}
      & $2$ & $-0.07\;[-0.37,\,0.24]$ & $14.3\%$ & $+0.41\;[0.07,\,0.75]$ & $28.6\%$ \\
      & $3$ & $+0.03\;[-0.28,\,0.33]$ & $4.2\%$ & $+0.51\;[0.16,\,0.86]$ & $43.8\%$ \\
    \addlinespace[2pt]
    TB-Dev & $3$ & $-0.59\;[-6.16,\,4.97]$ & $21.6\%$ & $+7.76\;[0.55,\,15.00]$ & $11.0\%$ \\
    \addlinespace[2pt]
    \multirow{2}{*}{ALFWorld}
      & $3$ & $-3.82\;[-7.00,\,-0.75]$ & $29.8\%$ & $+0.75\;[-2.15,\,3.73]$ & $18.3\%$ \\
      & $5$ & $-1.77\;[-4.57,\,0.93]$ & $22.0\%$ & $+2.80\;[-0.47,\,6.16]$ & $31.3\%$ \\
    \bottomrule
  \end{tabular}
\end{table}

\paragraph{A few teacher steps can approach the performance of full continuations.}
On HotpotQA, \OPshort{} with $\Ki=2$ or $3$ differs from \OPfull{} in test accuracy by only $-0.07$ or $+0.03$ percentage points (pp), while reducing the teacher-inference cost by $14.3\%$ or $4.2\%$, respectively.
On TB-Dev, \OPshort{} with $\Ki=3$ reduces teacher-inference cost by $21.6\%$ with an observed score difference of $-0.59$ pp, although the CI is wider.
On ALFWorld, \OPshort{} with $\Ki=5$ uses $22.0\%$ fewer teacher output tokens than \OPfull{}, with a test-accuracy difference of $-1.77$ pp.
Reducing the step cap to $\Ki=3$ gives a larger test-accuracy difference of $-3.82$ pp but with further cost reduction.
Therefore, we conclude that a short teacher continuation can provide a favorable cost-performance tradeoff, though the concrete step cap choice still depends on the environment.

\paragraph{A few teacher steps can steadily improve on one-step supervision.}
On HotpotQA, increasing $\Ki$ from $1$ to $2$ or $3$ improves test accuracy by $0.41$ or $0.51$ pp, respectively, with both CIs above zero.
On TB-Dev, increasing $\Ki$ from $1$ to $3$ improves the test score by $7.76$ pp, again with the CI above zero, at $11.0\%$ higher teacher-inference cost.
ALFWorld shows the same ordering: \OPshort{} with $\Ki=3$ and $5$ has $0.75$ and $2.80$ pp higher test accuracy than \OPshort{} with $\Ki=1$, at $18.3\%$ and $31.3\%$ higher teacher-inference cost.
All these results well motivate a few teacher steps as an intermediate choice between next-step-only supervision and teacher continuation to completion.

\subsection{Do filtering and truncation improve efficiency?}
\label{sec:q3-filtering-truncation}

To answer Q3, we study \OPsuccess{} and \OPcritical{} (Table~\ref{tab:design-space}), which we construct as ($\PhiPre, \PhiPost$)-filtered subsets of \OPfull{}, optionally truncating each filtered teacher continuation to only the first $\Ktr$ steps.
Table~\ref{tab:q3-filtering-truncation} changes one factor at a time.
Panel~(a) isolates filtering by comparing each recipe at $\Ktr=\Tmax$ against \OPfull{}, which differs only in ($\PhiPre, \PhiPost$).
Panel~(b) isolates truncation by comparing each recipe at a shorter $\Ktr$ against itself at $\Ktr=\Tmax$.
We omit TB-Dev because outcome verifiers for its training data are unavailable.

\begin{table}[htbp]
  \centering
  \caption{\emph{Filtering and truncation.}
  Mean accuracy paired differences [95\% CIs], in pp.
  Each cost reduction is relative to \OPfull{} in (a) and to the same recipe at $\Ktr=\Tmax$ in (b).}
  \label{tab:q3-filtering-truncation}
  \small
  \setlength{\tabcolsep}{3.5pt}
  \renewcommand{\arraystretch}{1.05}
  \begin{tabular*}{\linewidth}{@{\extracolsep{\fill}}llcrr@{}}
    \toprule
    \multicolumn{5}{@{}l}{\textbf{(a) Filtering:} each recipe at $\Ktr=\Tmax$ versus \OPfull{}} \\
    \midrule
    Benchmark & Recipe & Difference [95\% CI] & $\Ci$ reduction & $\Cg$ reduction \\
    \midrule
    \multirow{2}{*}{HotpotQA}
      & \OPsuccess{} & $-0.25\;[-0.57,\,0.07]$ & $0.0\%$ & $32.5\%$ \\
      & \OPcritical{} & $-0.48\;[-0.83,\,-0.13]$ & $30.4\%$ & $56.7\%$ \\
    \addlinespace[2pt]
    \multirow{2}{*}{ALFWorld}
      & \OPsuccess{} & $+1.21\;[-1.59,\,4.01]$ & $0.0\%$ & $10.8\%$ \\
      & \OPcritical{} & $-1.40\;[-4.29,\,1.40]$ & $41.5\%$ & $46.1\%$ \\
    \bottomrule
  \end{tabular*}%

  \vspace{3pt}
  \begin{tabular*}{\linewidth}{@{\extracolsep{\fill}}lccrcr@{}}
    \toprule
    \multicolumn{6}{@{}l}{\textbf{(b) Truncation:} each recipe at $\Ktr$ vs the same recipe (\OPsuccess{} / \OPcritical{}) without truncation.} \\
    \midrule
    & & \multicolumn{2}{c}{\OPsuccess{}} & \multicolumn{2}{c}{\OPcritical{}} \\
    \cmidrule(lr){3-4}\cmidrule(l){5-6}
    Benchmark & $\Ktr$ & Difference [95\% CI] & $\Cg$ reduction & Difference [95\% CI] & $\Cg$ reduction \\
    \midrule
    \multirow{3}{*}{HotpotQA}
      & $1$ & $-0.34\;[-0.66,\,-0.01]$ & $27.5\%$ & $-0.36\;[-0.69,\,-0.03]$ & $16.6\%$ \\
      & $2$ & $+0.02\;[-0.29,\,0.32]$ & $9.4\%$ & $+0.17\;[-0.15,\,0.50]$ & $6.5\%$ \\
      & $3$ & $+0.07\;[-0.23,\,0.38]$ & $1.8\%$ & $+0.34\;[0.02,\,0.66]$ & $1.6\%$ \\
    \addlinespace[2pt]
    \multirow{3}{*}{ALFWorld}
      & $1$ & $-2.80\;[-5.97,\,0.28]$ & $34.2\%$ & $-1.49\;[-4.38,\,1.40]$ & $3.8\%$ \\
      & $3$ & $-0.75\;[-3.17,\,1.77]$ & $23.0\%$ & $-0.09\;[-2.80,\,2.61]$ & $2.9\%$ \\
      & $5$ & $-1.96\;[-4.48,\,0.56]$ & $15.3\%$ & $+1.40\;[-0.93,\,3.73]$ & $2.2\%$ \\
    \bottomrule
  \end{tabular*}%
\end{table}

\paragraph{Outcome filtering has mixed effects on accuracy.}
Without truncation, \OPsuccess{} is slightly below \OPfull{} on HotpotQA and slightly above it on ALFWorld, with both CIs containing zero.
We thus cannot conclude that discarding failed teacher continuations helps, echoing findings that less aggressively filtered pretraining data can outperform quality-filtered data given enough compute \citep{goyal2024scaling,mohri2026bitter}.
For \OPsuccess{}, the clearer benefit is training cost, with $32.5\%$ and $10.8\%$ fewer retained tokens than \OPfull{} at the same $\Ci$.
\OPcritical{} continues only failed student rollouts and thus lowers both costs, $\Ci$ by $30.4\%$ and $41.5\%$ and $\Cg$ by roughly half, although its untruncated HotpotQA accuracy is $0.48$ pp below \OPfull{} (CI below zero).

\paragraph{Learning teacher continuations' long tail can be harmful.}
For \OPcritical{}, dicarding teacher steps after $\Ktr=3$ on HotpotQA and $\Ktr=5$ on ALFWorld removes only $1.6\%$ and $2.2\%$ of retained tokens, yet raises accuracy by $0.34$ pp (CI above zero) and $1.40$ pp.
We suspect that long continuations often come from a teacher that struggles before eventually succeeding, so their late steps, also the furthest from the student prefix, are poor imitation targets.

\paragraph{Truncation and filtering should be tuned jointly.}
The best $\Ktr$ depends on the filter: on ALFWorld, \OPsuccess{} is most accurate at full retention whereas \OPcritical{} peaks at $\Ktr=5$, and on HotpotQA a single retained step lowers accuracy for both recipes.
At its best observed $\Ktr$, \OPcritical{} is within $0.14$ pp of \OPfull{} on both benchmarks (CIs containing zero) while retaining $57.4\%$ and $47.3\%$ fewer tokens.

\section{Ablation Study}\label{sec:ablation}
Having compared recipes for expanding SFT data, we next ask whether the extra post-training budget to improve the \pureBC{}-coldstart student $\pi_1$ is better spent on non-SFT techniques such as OPD \citep{zhong2026vlaopd,agrawal2026distil} or GRPO-style RL \citep{deepseekmath}. We run two sets of comparison: (1). \OPshort{} against reverse-KL OPD on ALFWorld, where our local teacher hosting provides logits, and (2) \OPshort{} against GRPO and Dr.GRPO \citep{drgrpo} on HotpotQA.

\begin{center}
\begin{minipage}{\linewidth}
\centering
\includegraphics[width=\linewidth]{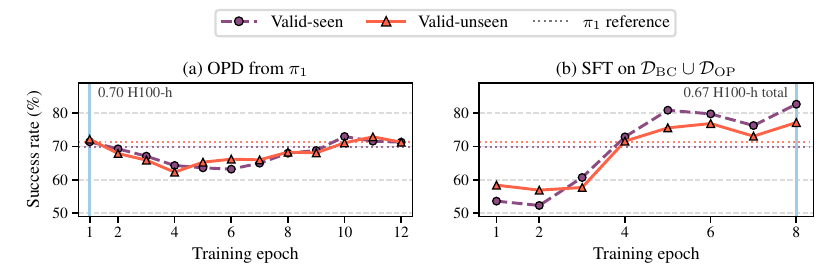}
\captionsetup{type=figure,font=small,labelfont=bf,skip=3pt,singlelinecheck=false}
\caption{\textbf{ALFWorld learning curves.} Dotted lines mark $\pi_1$
(69.8\% seen, 71.5\% unseen). OPD starts from $\pi_1$, while \OPshort{}
retrains the base student on BC+OP data. Labels give active H100-hours,
including collection and SFT for \OPshort{}.}
\label{fig:ablation-h100-epochs}
\end{minipage}
\end{center}

\textbf{OPD at higher GPU cost.}
We multiply active wall-clock time by GPU count, excluding pipeline startup, checkpoint
saving, and teardown for both methods.
In a separate 128-task ALFWorld ablation, \OPshort{} ($\Ki=3$) uses
0.67 H100-hours for student rollouts, teacher continuations, additional BC
demonstrations, and SFT altogether, versus 0.70 for one OPD epoch.
Its seen / unseen success rates are 82.7 / 77.2\%, compared with
71.4 / 72.2\% for OPD (Figure~\ref{fig:ablation-h100-epochs}).
Even across 12 epochs (9.9 H100-hours), OPD's best scores on the respective
splits are 73.0 / 72.9\%.

\textbf{RL compute to match OP-Short performance.}
On 3,000 HotpotQA dev questions, \OPshort{} ($\Ki=1$) reaches 50.4\% accuracy
at about 0.37 node-hour equivalents ($8\times$H100), including student rollouts,
teacher API queries, and the additional SFT run. We convert API expense to
equivalent rental time at \$20 per node-hour. RL costs include rollout generation,
log-probability computation, and policy updates, excluding validation and
checkpointing. From $\pi_1$, the first evaluated GRPO and Dr.GRPO
checkpoints exceeding 50.4\% require 1.67 and 2.09 node-hours of RL training,
respectively.

We provide experiment and cost-accounting details in Appendix~\ref{app:ablation-details}, including visualization of the RL runs (Figure~\ref{fig:ablation-rl}).

\section{Discussion}\label{sec:disc}

Once initial \pureBC{} produces a capable student $\pi_1$, teacher continuations from student-visited contexts expand an SFT dataset more cost-effectively than additional full demonstrations. Within this regime, the generation cap $\Ki$ governs a tradeoff between state coverage and per-turn cost: continuing for a few turns teaches local recovery and amortizes prompt-prefill caching beyond $\Ki=1$, without spending $\Ci$ on late teacher-shaped states as in $\Ki=\Tmax$. Because the best $\Ki$ varies by environment, continuation length remains task-dependent. Filtering likewise depends on which cost dominates: post-filtering ($\PhiPost$) cuts retained tokens $\Cg$ even though discarded rollouts still add to $\Ci$, whereas pre-filtering failed student rollouts (\OPcritical{}) directs queries to unresolved errors and so could be more cost-effective. SFT on brief continuations also works with black-box teachers and matches or exceeds RL and reverse-KL OPD at much lower compute.

\paragraph{Limitations.}
We study single-round expansion from a fixed $\pi_1$ across three benchmarks, with filtering sweeps on HotpotQA and ALFWorld. Relative gains may shift under multi-round collection or when outcomes cannot be automatically verified for \OPcritical{}. API dollar figures use reconstructed cache-hit rates. We also lack a theoretical rule for choosing $\Ki$ without tuning.

\clearpage
\section*{Acknowledgement}

We thank Stanford Marlowe Cluster \citep{kapfer2025marlowe} for providing GPU compute. Mohsen Bayati and Junze Ye thank the Stanford Graduate School of Business for providing research grants.
Jose Blanchet gratefully acknowledges supports from ONR under award N00014-24-1-2655, and the National Science Foundation (NSF) under grants 2312204 and 2403007. All of the authors would like to thank the anonymous NeurIPS reviewers for constructive feedback during the rebuttal process.
\bibliographystyle{plainnat}
\bibliography{references}

\clearpage
\appendix
\section{Related Work}\label{sec:literature}
\paragraph{Interactive imitation learning.}
The \textsc{OP} data recipes we study are inspired by DAgger
\citep{smile, ross2011dagger, ross2014}, which mitigates train-test distribution shift by collecting teacher labels at student-induced states and adding them to the training data. Since querying the expert at every visited state is costly, later variants restrict when the expert acts. HG-DAgger \citep{hgdagger} lets a human expert decide when to take over, whereas SafeDAgger \citep{safedagger}, LazyDAgger \citep{lazydagger}, and ThriftyDAgger \citep{thriftydagger} learn a gating rule that triggers expert takeovers, with the latter two also limiting how often control changes hands. These methods target continuous-control tasks such as driving and robotics, typically with a human supervisor whose cost is counted in actions and interventions. We instead study multi-turn LLM agents taught by a larger LLM, and measure teacher-inference cost at the token level. How performance scales with the supervision budget has also received less attention. \citet{thriftydagger} evaluate a single target intervention rate and leave the dependence on supervisor burden to future work, whereas our experiments trace student performance against a common teacher-cost axis.

\paragraph{DAgger-style LLM agent finetuning.}
\citet{caccia2024finetuningwebagents}
show that, for LLM agents, better prediction on expert trajectories need
not translate to better downstream performance once the student acts on
its own induced contexts, and mention DAgger as a potential remedy. \citet{lauffer2025oec} are the closest prior
work to our intervention-style setup: they roll in with the student to a
random switch step, let the expert continue to termination, mask the
student prefix during training, and filter by verifiable success and
student repetition. This motivates our \OPsuccess{} baselines, whereas our
main \OPshort{} method keeps every sampled continuation but bounds it to a few teacher turns. Other variants use privileged train-time teacher
state \citep{choudhury2024better}, teacher-identified first-error
corrections \citep{lyu2025score}, self-reflected teacher-guided
corrections \citep{chen2025step}, or turn-level teacher-student mixing
for long-horizon software agents \citep{revisitDagger}. 

\paragraph{On-policy distillation.}
A related line studies on-policy distillation (OPD) for autoregressive
language models, training on student-sampled sequences while matching
teacher behavior via generalized knowledge distillation
\citep{agarwal2024gkd}, reverse-KL objectives \citep{gu2024minillm},
context internalization \citep{ye2026opcd}, privileged self-distillation
\citep{zhao2026opsd}, or verbal-score supervision \citep{xiong2026ovd}.
Concurrently, \citet{zhao2026decoupling} decouple prefix source from KL
direction, identifying student-prefix forward-KL as DAgger-style
on-policy SFT and student-prefix reverse-KL as OPD. In addition, \citet{he2026pivotopd} have the teacher identify pivotal mistakes in student rollouts and supervise the corrected action together
with a few recovery turns; \OPcritical{} shares the intuition of
concentrating supervision where the student goes wrong. Unlike most OPD
methods, our data recipes' SFT setup does not require access to the teacher's next-token probability distribution.

\paragraph{Offline-to-online learning.}
Online RL often needs a large number of environment interactions when it starts from a weak policy. A common remedy is to first learn from offline data and then fine-tune online. \citet{nakamoto2023calql} studies what makes an offline initialization amenable to fast online fine-tuning, and \citet{bai2024digirl} applies this two-stage paradigm to device-control agents. Our setting follows a similar progression. The \textsc{OP} recipes can be viewed as a ``mid-training'' stage between \pureBC{} and online RL, since the student learns on contexts induced by its own rollouts, as in online RL, but from teacher actions instead of rewards. A related family of methods learns from logged demonstrations by weighting rewards \citep{peters2007rwr} or advantage estimates \citep{peng2019awr, nair2020awac}. SFT on success-filtered trajectories is the special case of binary trajectory-level weights, which \citet{qin2025iwsft} show optimizes a lower bound on the sparse-reward RL objective. The filters and switch-time distribution of our \textsc{OP} recipes act as a coarse turn-level weighting in a similar spirit.

\paragraph{Allocating expert effort.}
Deciding where to spend scarce expert effort is a recurring question across many research communities. Active learning selects which unlabeled examples to query \citep{settles2009active}, while dataset audits and benchmark stewardship direct expert review toward likely errors \citep{northcutt2021pervasive, ye2025stewardship}. Learning to defer routes individual decisions between a model and a human expert \citep{madras2018defer, mozannar2020consistent}, and work on human-AI teams studies when costly human input is worth acquiring \citep{raghu2019triage, wilder2020complement, tan2026dispatch}. Our work studies an analogous allocation problem in SFT data collection for multi-turn LLM agents, where the scarce expert is a stronger teacher model.

\paragraph{Positioning relative to prior work.}
Recent agent post-training recipes often bundle several choices: prefix
policy, switch-time sampling, continuation length, masking, filtering,
and sometimes a subsequent RL stage. We instead make these choices
explicit and compare them under matched supervision resources, separating
teacher-inference cost from retained SFT training cost. This isolates the
question central to our experiments: once a reasonable behavioral-cloning
student is available, whether additional teacher budget is better spent
on more start-to-end expert demonstrations, longer or filtered teacher
continuations, or short unfiltered corrections spread over more
learner-induced contexts.
\section{Mathematical Formulation}
\label{app:math-details}

\S\ref{sec:context-sft-dist} states that each recipe $\mathcal G=(\Tswitch,\PhiPre,\PhiPost,\Ki,\Ktr,n)$ induces its own distribution $d$ over SFT contexts. This appendix formalizes that distribution and shows how every data-collection choice enters it as a multiplicative weight. The switch-time rule $\Tswitch$ and the filters ($\PhiPre, \PhiPost$) weight the student steps\footnote{``Step'' and ``turn'' mean the same thing in this paper. The main text uses both, while this appendix and Appendix~\ref{app:switch-time} try to be more consistent and use ``step'' throughout.} at which the teacher takes over (\S\ref{app:context-distribution}). In addition, the step caps ($\Ki, \Ktr$) and the post-filter $\PhiPost$ determine how many and which teacher-continued contexts become training data (\S\ref{app:multiturn-sft}), and we generalize the SFT objective (Eq.~\ref{eq:population-sft-ce}) to multi-step teacher continuations. Appendix~\ref{app:switch-time} compares our $\Tswitch$ with that of OEC \citep{lauffer2025oec}.

\paragraph{Conventions.}
We assume that tasks, agent actions, and environment responses are finite strings, so we could represent every distribution as a probability mass function and every expectation as a sum. Let $p^{\pi}(\tau)$ be the probability that drawing a task $q\sim\mu$ and rolling out the student $\pi$ to termination produces the trajectory $\tau$. By Eq.~\ref{eq:pomdp-dynamics},
\begin{equation}
p^{\pi}(\tau)=\mu(q)\prod_{t=1}^{T(\tau)}
\pi(a_t\mid h_{t-1})\,\mathrm{P}_s(s_t\mid s_{t-1},a_t)\,\mathrm{P}_o(o_t\mid s_t,a_t),
\label{eq:rollout-pmf}
\end{equation}
and $\mathbb E_{\tau}$ denotes an expectation over $\tau\sim p^{\pi}$. A \emph{student step} is a pair $(\tau,t)$ with $1\le t\le T(\tau)$, and its LLM input context is $h_{t-1}(\tau)$. We additionally denote $p^{\pie}(\tau'\mid\tau,t')$ as the probability that letting the teacher $\pie$ branch from timestep $t'$ of $\tau$ for at most $\Ki$ steps produces the hybrid trajectory $\tau'$ (\S\ref{sec:methods-data}). this has a similar product form as Eq.~\ref{eq:rollout-pmf}. We write $k_e(\tau')$, or simply $k_e$, for the number of teacher steps in $\tau'$. Filters are $\{0,1\}$-valued functions, and $\Tswitch(t'\mid\tau)$ is the probability of proposing switch time $t'$ given $\tau$. One \emph{proposal} involves first drawing $\tau\sim p^{\pi}$, then $t'\sim\Tswitch(\cdot\mid\tau)$, and finally $\tau'\sim p^{\pie}(\cdot\mid\tau,t')$. $\mathbb E_{\tau,t',\tau'}$ averages over all three sequential draws, while $\mathbb E_{\tau'}$ marginalizes over the teacher continuation alone with $(\tau,t')$ held as given.

\subsection{Data collection procedure}
\label{app:data-collection-algorithm}

\begin{algorithm}[h]
\caption{Constructing SFT data from collection hyperparameters $\mathcal G$.}
\label{alg:sft-rollout}
\small
\begin{algorithmic}[1]
\Require Tasks $\Qtrain$, fixed policies $\pi,\pie$, data recipe $\mathcal G=(\Tswitch,\PhiPre,\PhiPost,\Ki,\Ktr,n)$.
\State $\DG\gets[\,]$, \quad $(\Ci,\Cg)\gets(0,0)$
\For{$q\in\Qtrain$}
  \For{$\ell=1,\ldots,n$} \Comment{Each iteration is one \emph{proposal}}
    \If{the recipe is \pureBC}
      \State $t'\gets1$; \ $(h,s)\gets$ initial context and environment state of $q$
    \Else
      \State $\tau\gets$ a completed rollout of $\pi$ on $q$ \Comment{$p^{\pi}(\tau)$}
      \State Sample $t'\sim\Tswitch(\cdot\mid\tau)$ \Comment{$\Tswitch(t'\mid\tau)$}
      \If{$t'>T(\tau)$ or $\PhiPre(\tau,t')=0$} \Comment{$\ind{t'\le T(\tau)}\,\PhiPre(\tau,t')$}
        \State \textbf{continue} \Comment{no teacher call}
      \EndIf
      \State $(h,s)\gets(h_{t'-1},s_{t'-1})$ \Comment{context and environment state of $\tau$ before step $t'$}
    \EndIf
    \State Roll out $\pie$ from $(h,s)$ for at most $\min\{\Ki,\Tmax-t'+1\}$ steps,
    \Statex \hspace{\algorithmicindent}\hspace{\algorithmicindent}\hspace{\algorithmicindent}producing $\tau'$ with $k_e$ teacher steps \Comment{$p^{\pie}(\tau'\mid\tau,t')$}
    \State Add the teacher-inference cost of all $k_e$ steps to $\Ci$
    \If{$\PhiPost(\tau')=1$} \Comment{$\PhiPost(\tau')$}
      \State $m\gets\min\{k_e,\Ktr\}$
      \State Append $\bigl(h,(a^\star_{t'},o'_{t'},\ldots,a^\star_{t'+m-1},o'_{t'+m-1})\bigr)$ to $\DG$
      \State $\Cg\gets\Cg+$ number of tokens in $a^\star_{t'},\ldots,a^\star_{t'+m-1}$
    \EndIf
  \EndFor
\EndFor
\State \Return $\DG$, $\Ci$, $\Cg$
\end{algorithmic}
\end{algorithm}

Algorithm~\ref{alg:sft-rollout} is the proposal process of \S\ref{sec:methods-data}. Some comments on the right name the probability factor that the corresponding line contributes, and \S\ref{app:context-distribution} shows how they multiply together. The student rollout $\tau$ is completed before a switch time is drawn, so $\PhiPre(\tau,t')$ may inspect the student's terminal outcome, as in \OPcritical{}. The teacher stops before the cap whenever it completes the task, so $k_e$ can be smaller than $\min\{\Ki,\Tmax-t'+1\}$. In our paper, every task receives $n$ proposals, so $\mu$ is uniform over $\Qtrain$.

\subsection{Recipes as weights on student steps}
\label{app:context-distribution}

\paragraph{The weight of a student step.}
Fix an \textsc{OP} recipe and consider one proposal. Reading off the comments in Algorithm~\ref{alg:sft-rollout}, the proposal is accepted with student rollout $\tau$ and switch time $t\le T(\tau)$ with probability
\begin{equation}
p^{\pi}(\tau)\,w_t(\tau),
\qquad
w_t(\tau):=
\underbrace{\Tswitch(t\mid\tau)}_{\text{switch time}}\cdot
\underbrace{\PhiPre(\tau,t)}_{\text{pre-filter}}\cdot
\underbrace{\mathbb E_{\tau'}[\PhiPost(\tau')]}_{\text{post-filter}},
\label{eq:step-weight}
\end{equation}
where the last factor is the probability that a teacher continuation from step $t$ passes $\PhiPost$. Summing over all student steps gives the acceptance rate of a proposal, $Z_{\mathcal G}:=\mathbb E_{\tau}\bigl[\sum_{t=1}^{T(\tau)}w_t(\tau)\bigr]$, a normalization constant. The expected number of collected examples is $n\,|\Qtrain|\,Z_{\mathcal G}$. The distribution of the starting context $h$ of an accepted proposal can be described by a probability mass function
\begin{equation}
d_{\mathcal G}(h):=\mathrm{Pr}(\mathbf{h} = h)
=\frac{1}{Z_{\mathcal G}}\,
\mathbb E_{\tau}\!\left[\sum_{t=1}^{T(\tau)}w_t(\tau)\,\ind{h_{t-1}(\tau)=h}\right],
\label{eq:general-occupancy}
\end{equation}
where the bolded $\mathbf{h}$ denotes the random variable version of $h$. This is the context distribution $d$ of \S\ref{sec:context-sft-dist} for the first teacher step.

\paragraph{Task weights.}
The remaining factor $\mu(q)$ in $p^{\pi}(\tau)$ (Eq.~\ref{eq:rollout-pmf}) is a weight on tasks rather than on steps. Task-curation methods change $\mu$. In RL post-training, for example, DAPO \citep{yu2025dapo} and online difficulty filtering \citep{bae2025online} estimate each prompt's success rate under the current policy from sampled rollouts and keep task prompts whose success rate is neither too low nor too high, while prompt curriculum learning \citep{gao2026pcl} predicts this success rate with a learned value model. Our recipes instead hold $\mu$ fixed (uniform over $\Qtrain$) and change only $w_t(\tau)$, so the two choices are complementary and their weights multiply.

\paragraph{Importance-ratio form.}
A natural starting point is $w_t(\tau)\equiv1$, which weights all student steps equally. Eq.~\ref{eq:general-occupancy} then simplifies to:
\begin{equation}
d_{\mathrm{step}}(h)
:=\frac{\mathbb E_{\tau}\bigl[\sum_{t=1}^{T(\tau)}\ind{h_{t-1}(\tau)=h}\bigr]}{\mathbb E_{\tau}[T(\tau)]}.
\label{eq:pooled-occupancy}
\end{equation}
This \emph{per-step} distribution is what one obtains by putting the steps of many student trajectories into one bag and drawing a step uniformly at random. Comparing Eq.~\ref{eq:general-occupancy} with Eq.~\ref{eq:pooled-occupancy},
\begin{equation}
d_{\mathcal G}(h)
=\frac{\mathbb E_{\tau}[T(\tau)]}{Z_{\mathcal G}}\;\bar w_{\mathcal G}(h)\,d_{\mathrm{step}}(h),
\qquad
\bar w_{\mathcal G}(h):=
\frac{\mathbb E_{\tau}\bigl[\sum_{t}w_t(\tau)\,\ind{h_{t-1}(\tau)=h}\bigr]}
     {\mathbb E_{\tau}\bigl[\sum_{t}\ind{h_{t-1}(\tau)=h}\bigr]},
\label{eq:importance-ratio}
\end{equation}
where $\bar w_{\mathcal G}(h)$ is the average weight of the student steps whose context is $h$, and $\mathbb E_{\tau}[T(\tau)]/Z_{\mathcal G}$ is a normalizing constant. Hence a data recipe $\mathcal G$ is an importance reweighting of the per-step distribution, which acts as the reference distribution, and the mass of each context $h$ is rescaled in proportion to $\bar w_{\mathcal G}(h)$. Write $\mathrm{CE}_\theta(h):=\mathbb E_{a^\star\sim\pie(\cdot\mid h)}[-\log\pi_\theta(a^\star\mid h)]$ for the per-context loss in Eq.~\ref{eq:population-sft-ce}. The importance-sampling identity $\mathbb E_{h\sim d_{\mathcal G}}[f(h)]=\mathbb E_{h\sim d_{\mathrm{step}}}\bigl[\tfrac{d_{\mathcal G}(h)}{d_{\mathrm{step}}(h)}f(h)\bigr]$ then gives
\begin{equation}
\mathcal L_{\mathrm{SFT}}(\theta;d_{\mathcal G})
=\mathbb E_{h\sim d_{\mathcal G}}\bigl[\mathrm{CE}_\theta(h)\bigr]
=\frac{\mathbb E_{\tau}[T(\tau)]}{Z_{\mathcal G}}\;
\mathbb E_{h\sim d_{\mathrm{step}}}\bigl[\bar w_{\mathcal G}(h)\,\mathrm{CE}_\theta(h)\bigr].
\label{eq:weighted-sft}
\end{equation}

\paragraph{Weights of the recipes in Table~\ref{tab:design-space}.}
Table~\ref{tab:recipe-weights} lists the three factors of Eq.~\ref{eq:step-weight}. Each has a plain reading. The switch-time rule decides how a rollout's mass is spread over its steps. The pre-filter zeroes out student steps using student-side information only; \OPcritical{} removes every step of a successful student rollout. We abbreviate the post-filter factor as
\begin{equation}
\alpha(\tau,t):=\mathbb E_{\tau'}[\PhiPost(\tau')],
\label{eq:post-filter-factor}
\end{equation}
which is a soft weight in $[0,1]$. When $\PhiPost$ is a success-only filter, as in \OPsuccess{} and \OPcritical{}, contexts from which the teacher rarely succeeds are down-weighted in proportion to its success probability.

\begin{table}[H]
  \centering
  \caption{Weight $w_t(\tau)$ that each recipe places on student step $(\tau,t)$, up to a constant. For OEC, $T_0$ is the hand-picked constant, and $\alpha$ accounts for its action-repetition filter. For LEAP/SCoRe, $J(t\mid\tau)$ is the probability that the teacher judge selects step $t$.}
  \label{tab:recipe-weights}
  \small
  \renewcommand{\arraystretch}{1.2}
  \vspace{4pt}
  \begin{tabular}{lcccc}
  \toprule
  \textbf{Recipe} & $\Tswitch(t\mid\tau)$ & $\PhiPre(\tau,t)$ & $\alpha(\tau,t)$ & $w_t(\tau)\ \propto$ \\
  \midrule
  \OPshort, \OPfull & $1/T(\tau)$ & $1$ & $1$ & $1/T(\tau)$ \\
  \OPsuccess & $1/T(\tau)$ & $1$ & $\Pr[\text{teacher succeeds}]$ & $\alpha(\tau,t)/T(\tau)$ \\
  \OPcritical & $1/T(\tau)$ & $\ind{\tau\text{ fails}}$ & $\Pr[\text{teacher succeeds}]$ & $\ind{\tau\text{ fails}}\,\alpha(\tau,t)/T(\tau)$ \\
  \midrule
  OEC & $1/T_0$ & $1$ & $\Pr[\text{teacher succeeds}]$ & $\alpha(\tau,t)$ \\
  LEAP/SCoRe & $J(t\mid\tau)$ & $\ind{\tau\text{ fails}}$ & $1$ & $\ind{\tau\text{ fails}}\,J(t\mid\tau)$ \\
  \bottomrule
  \end{tabular}
\end{table}

\paragraph{Weights versus cost.}
The three factors change the context distribution in the same multiplicative way, but they are paid for differently. An invalid switch time or a rejection by $\PhiPre$ happens before any teacher call. A rejection by $\PhiPost$ happens after the teacher has generated all $k_e(\tau')$ steps. Let $A_{\mathrm{pre}}:=\ind{t'\le T(\tau)}\,\PhiPre(\tau,t')$ indicate that a student context proposal reaches the teacher. Per proposal, the expected numbers of teacher steps generated and retained are
\begin{equation}
\text{generated:}\ \ \mathbb E_{\tau,t',\tau'}\bigl[A_{\mathrm{pre}}\,k_e(\tau')\bigr],
\qquad
\text{retained:}\ \ \mathbb E_{\tau,t',\tau'}\bigl[A_{\mathrm{pre}}\,\PhiPost(\tau')\,\min\{k_e(\tau'),\Ktr\}\bigr].
\label{eq:expected-steps}
\end{equation}
The costs $\Ci$ and $\Cg$ of \S\ref{sec:cost-model} are the dollar- and token-denominated versions of these two quantities. $\PhiPost$ and $\Ktr$ appear only in the second expression, which is why they lower $\Cg$ without lowering $\Ci$.

\subsection{Multi-step continuations and the SFT objective}
\label{app:multiturn-sft}

Eq.~\ref{eq:population-sft-ce} assumes one teacher continuation step $a^\star_{t}$ per context $h_{t-1}$. In the case where the teacher ($\pi_e$) continues for multiple steps and we retain $m(\tau')=\min\{k_e(\tau'),\Ktr\}$ of these teacher steps, the training loss becomes
\begin{equation}
\ell_\theta(\tau',t')
:=\sum_{j=0}^{m(\tau')-1}-\log\pi_\theta\bigl(a^\star_{t'+j}\mid h_{t'+j-1}(\tau')\bigr).
\label{eq:multi-step-loss}
\end{equation}
The first context $h_{t'-1}(\tau')=h_{t'-1}(\tau)$ is the student's ($\pi_{\theta}$). Each later context appends the preceding teacher steps and environment responses, which enter only as conditioning text (teacher forcing). We call these later contexts \emph{teacher-continued} contexts.

\paragraph{From $d_{\mathcal G}$ to a mixture distribution.}
Eq.~\ref{eq:multi-step-loss} is a sum of $m(\tau')$ single-step losses, one for each context $h_{t'+j-1}(\tau')$ at which the student is made to imitate a teacher step. Let $d'_{\mathcal G}$ be the distribution of these contexts, obtained by putting the contexts of all accepted proposals into one bag and drawing one uniformly at random. With $\mathbb E$ denoting an average over accepted SFT data proposals throughout this subsection,
\begin{equation}
d'_{\mathcal G}(h)
:=\frac{1}{\bar m_{\mathcal G}}\,
\mathbb E\Bigl[\,\sum_{j=0}^{m(\tau')-1}\ind{h_{t'+j-1}(\tau')=h}\Bigr],
\label{eq:mixture-context}
\end{equation}
where $\bar m_{\mathcal G}:=\mathbb E[m(\tau')]$ is the average number of retained teacher steps per accepted proposal. When $\PhiPost$ accepts everything (\pureBC{}, \OPshort{}, \OPfull{}), the average of Eq.~\ref{eq:multi-step-loss} is
\begin{equation}
\mathbb E\bigl[\ell_\theta(\tau',t')\bigr]
=\bar m_{\mathcal G}\cdot\mathcal L_{\mathrm{SFT}}(\theta;d'_{\mathcal G}),
\label{eq:multi-step-reduction}
\end{equation}
so multi-step SFT is the single-step objective of Eq.~\ref{eq:population-sft-ce} with $d'_{\mathcal G}$ in place of $d$, up to the constant $\bar m_{\mathcal G}$.

This $d'_{\mathcal G}$ is a mixture: a fraction $1/\bar m_{\mathcal G}$ of its mass sits on student contexts drawn from $d_{\mathcal G}$, and the rest sits on teacher-continued contexts. With $\Ki=1$ or $\Ktr=1$, every accepted proposal retains one teacher step and $d'_{\mathcal G}=d_{\mathcal G}$. Larger caps move mass toward teacher-continued contexts, trading coverage of distinct student contexts for longer teacher continuations that are rarer (i.e. more off-policy) for the student to encounter. \pureBC{} is the extreme case: $t'=1$ and the whole teacher rollout is retained, so $d'_{\mathcal G}$ is the per-step distribution of Eq.~\ref{eq:pooled-occupancy} with $\pie$ in place of $\pi$.

\paragraph{With a post-filter.}
For \OPsuccess{} and \OPcritical{}, $\PhiPost$ depends on the teacher's own later steps. Among accepted proposals, a continuation from student step $(\tau,t)$ has probability
\begin{equation}
p^{\pie}(\tau'\mid\tau,t)\cdot\frac{\PhiPost(\tau')}{\alpha(\tau,t)},
\label{eq:filtered-continuation}
\end{equation}
which is again an importance reweighting, now of teacher continuations, with ratio $\PhiPost/\alpha$. The retained steps are draws from the teacher conditioned on its continuation being accepted, rather than fresh draws from $\pie(\cdot\mid h)$. Eq.~\ref{eq:multi-step-reduction} then holds with $\pie$ replaced by the conditional distribution of retained steps given their context, i.e., a success-conditioned teacher. 

\section{Sampling the Switch Time}
\label{app:switch-time}

This appendix compares our switch-time rule with that of OEC \citep{lauffer2025oec} (\S\ref{app:pooled-switch-equivalence}) and gives an implementation of ours that needs fewer rollout playbacks (\S\ref{app:lazy-switch}). It uses the notation of Appendix~\ref{app:math-details}.

\subsection{Our switch-time rule versus OEC's}
\label{app:pooled-switch-equivalence}

To isolate the switch-time rule, set both filters to always passing so that $w_t(\tau)=\Tswitch(t\mid\tau)$. Our rule and OEC's both draw the switch time uniformly. They differ in the unit that receives equal weight: ours gives every student \emph{trajectory} the same total mass (per-trajectory weighting), whereas OEC's gives every student \emph{step} the same mass (per-step weighting).

\paragraph{Ours: every trajectory counts equally.}
We draw $t'\sim\mathrm{Unif}\{1,\ldots,T(\tau)\}$ after observing the completed trajectory, so $w_t(\tau)=1/T(\tau)$. Every $t'$ proposal is a valid step, and Eq.~\ref{eq:general-occupancy} becomes
\begin{equation}
d_{\mathrm{traj}}(h)
:=\mathbb E_{\tau}\!\left[\frac{1}{T(\tau)}\sum_{t=1}^{T(\tau)}\ind{h_{t-1}(\tau)=h}\right].
\label{eq:two-occupancies}
\end{equation}
Each trajectory contributes total mass $p^{\pi}(\tau)$, split evenly over its timesteps.

\paragraph{OEC: every step counts equally.}
OEC draws $t'\sim\mathrm{Unif}\{1,\ldots,T_0\}$ from a fixed horizon $T_0$, independently of $\tau$, and discards the proposal when $t'>T(\tau)$ (Table~\ref{tab:design-space}; $T_0=31$ in their experiments under array $1$-indexing, and we assume $T_0\ge \Tmax$). Every valid step then has the same weight $w_t(\tau)=1/T_0$. The constant cancels in Eq.~\ref{eq:general-occupancy}:
\begin{equation}
d_{\mathcal G}(h)
=\frac{\mathbb E_{\tau}\bigl[\sum_{t=1}^{T(\tau)}\tfrac{1}{T_0}\,\ind{h_{t-1}(\tau)=h}\bigr]}
      {\mathbb E_{\tau}[T(\tau)]/T_0}
=d_{\mathrm{step}}(h).
\label{eq:oec-pooled}
\end{equation}
So fixed-horizon sampling with invalid-switch rejection is exactly per-step weighting. A trajectory of length $T(\tau)$ survives the rejection with probability $T(\tau)/T_0$, and therefore contributes total mass proportional to $p^{\pi}(\tau)\,T(\tau)$.

\paragraph{Comparison.}
By Eq.~\ref{eq:step-weight}, the importance ratio of our rule to OEC's on a student step $(\tau,t)$ is
\begin{equation}
\frac{\text{mass under ours}}{\text{mass under OEC}}
=\frac{p^{\pi}(\tau)/T(\tau)}{p^{\pi}(\tau)/\mathbb E_{\tau}[T(\tau)]}
=\frac{\mathbb E_{\tau}[T(\tau)]}{T(\tau)}.
\label{eq:ours-vs-oec}
\end{equation}
Relative to OEC, our rule up-weights steps from shorter-than-average student trajectories and down-weights steps from longer ones. Under per-step weighting, a trajectory's share grows with its length, so long trajectories can crowd out short ones (Figure~\ref{fig:state-averaging}), and those that exhaust the step budget $\Tmax$ receive the largest share. The two rules also differ in yield, and they agree in one special case.
\begin{itemize}[leftmargin=1.5em,itemsep=1pt,topsep=2pt]
\item \emph{Yield.} Under our rule each proposal reaches the pre-filter, so with no filtering every task contributes exactly $n$ starting contexts. Under OEC's rule a fraction $1-\mathbb E_{\tau}[T(\tau)]/T_0$ of proposals is discarded before the teacher is queried.
\item \emph{When they agree.} If all trajectories have the same length, as in DAgger's fixed-horizon setting \citep{ross2011dagger}, then $T(\tau)=\mathbb E_{\tau}[T(\tau)]$ and the two distributions coincide.
\end{itemize}
The filters of \OPsuccess{} and \OPcritical{} multiply either rule's weight by the same factor $\PhiPre(\tau,t)\,\alpha(\tau,t)$, so the ratio in Eq.~\ref{eq:ours-vs-oec} remains proportional to $1/T(\tau)$.

\paragraph{Effect on task weights.}
By Eqs.~\ref{eq:two-occupancies} and~\ref{eq:pooled-occupancy}, the share of starting contexts that come from task $q$ is
\begin{equation}
\text{per-trajectory:}\ \ \mu(q),
\qquad\qquad
\text{per-step:}\ \ \mu(q)\,\frac{\mathbb E_{\tau}[T(\tau)\mid q]}{\mathbb E_{\tau}[T(\tau)]}.
\label{eq:task-share}
\end{equation}
Per-trajectory weighting keeps the task weights $\mu$, whereas per-step weighting rescales them by each task's expected trajectory length. In Figure~\ref{fig:state-averaging}, two equally likely task types end up with shares of $2\%$ and $98\%$.

One may object that tasks with long trajectories deserve this extra supervision. If they do, the place to say so is $\mu$ (\S\ref{app:context-distribution}, Task weights), which can be set independently of the switch-time rule instead of being fixed implicitly by trajectory length. Without such knowledge, keeping $\mu$ is the neutral choice, for two reasons. First, if test tasks are drawn from $\mu$, the training contexts cover tasks in the same proportions. Second, the steps of one trajectory share a task, an environment, and a prefix, so they add less diverse signal than the same number of steps spread over different trajectories.

\begin{figure}[H]
  \centering
  \begingroup
\definecolor{cwBlue}{HTML}{2A78D6}
\definecolor{cwOrange}{HTML}{EB6834}
\colorlet{cwBlueFill}{cwBlue!20!white}
\colorlet{cwOrangeFill}{cwOrange!22!white}
\colorlet{cwMuted}{black!62}
\begin{tikzpicture}[x=1cm,y=1cm,
  every node/.style={font=\footnotesize, inner sep=1.5pt, text=black},
  short/.style={draw=cwBlue, fill=cwBlueFill, line width=0.5pt},
  long/.style={draw=cwOrange, fill=cwOrangeFill, line width=0.5pt}]
\def\X{3.25}   %
\def\W{10.2}   %
\def\c{0.66}   %
\def\H{0.50}   %

\node[anchor=west, inner sep=0pt] at (0,5.55)
  {\textbf{Two task types} \textcolor{cwMuted}{(each with probability $\tfrac12$; one box per step of its student trajectory)}};

\node[anchor=west, inner sep=0pt, align=left] at (0,4.80)
  {task type 1 (short)\\[-1pt]\textcolor{cwMuted}{$\tau_1$,\ \ $T(\tau_1)=2$}};
\foreach \i in {0,1}
  \draw[short] (\X+\i*\c,4.55) rectangle ++(\c,\H) node[midway] {$h_{\i}$};

\node[anchor=west, inner sep=0pt, align=left] at (0,3.95)
  {task type 2 (long)\\[-1pt]\textcolor{cwMuted}{$\tau_2$,\ \ $T(\tau_2)=100$}};
\foreach \i in {0,...,5}
  \draw[long] (\X+\i*\c,3.70) rectangle ++(\c,\H) node[midway] {$h_{\i}$};
\foreach \i/\lab in {2/98,1/99}
  \draw[long] (\X+\W-\i*\c,3.70) rectangle ++(\c,\H) node[midway] {$h_{\lab}$};
\draw[cwOrange, line width=1.2pt, dash pattern=on 0pt off 6pt, line cap=round]
  (\X+6*\c+0.25,3.70+0.5*\H) -- (\X+\W-2*\c-0.2,3.70+0.5*\H);

\draw[black!25, line width=0.4pt] (0,3.15) -- (\X+\W,3.15);
\node[anchor=west, inner sep=0pt] at (0,2.72)
  {\textbf{Share of the teacher's starting contexts} \textcolor{cwMuted}{(box width $=$ probability of that step)}};

\pgfmathsetmacro{\u}{\W/102}
\node[anchor=west, inner sep=0pt, align=left] at (0,1.83)
  {per-step weighting\\[-1pt]\textcolor{cwMuted}{$d_{\mathrm{step}}$}};
\draw[short] (\X,1.58) rectangle ++(2*\u,\H);
\draw[long]  (\X+2*\u,1.58) rectangle (\X+\W,1.58+\H);
\draw[cwBlue, line width=0.3pt] (\X+\u,1.58) -- ++(0,\H);
\foreach \i in {3,...,101}
  \draw[cwOrange!70!white, line width=0.25pt] (\X+\i*\u,1.58) -- ++(0,\H);
\draw[cwOrange, line width=0.5pt] (\X+2*\u,1.58) rectangle (\X+\W,1.58+\H);
\node[anchor=north west, inner sep=0pt] at (\X-0.02,1.47)
  {type 1: $\tfrac{2}{102}\approx 2\%$};
\node[anchor=north, inner sep=0pt] at (\X+2*\u+50*\u,1.47)
  {type 2: $\tfrac{100}{102}\approx 98\%$ \textcolor{cwMuted}{(every step $\tfrac{1}{102}$)}};

\pgfmathsetmacro{\v}{\W/200}
\node[anchor=west, inner sep=0pt, align=left] at (0,0.45)
  {per-trajectory weighting\\[-1pt]\textcolor{cwMuted}{$d_{\mathrm{traj}}$}};
\foreach \i in {0,1}
  \draw[short] (\X+\i*0.25*\W,0.20) rectangle ++(0.25*\W,\H) node[midway] {$h_{\i}$};
\draw[long] (\X+0.5*\W,0.20) rectangle (\X+\W,0.20+\H);
\foreach \i in {101,...,199}
  \draw[cwOrange!70!white, line width=0.25pt] (\X+\i*\v,0.20) -- ++(0,\H);
\draw[cwOrange, line width=0.5pt] (\X+0.5*\W,0.20) rectangle (\X+\W,0.20+\H);
\node[anchor=north, inner sep=0pt] at (\X+0.25*\W,0.09)
  {type 1: $\tfrac12$ \textcolor{cwMuted}{(each step $\tfrac14$)}};
\node[anchor=north, inner sep=0pt] at (\X+0.75*\W,0.09)
  {type 2: $\tfrac12$ \textcolor{cwMuted}{(each step $\tfrac{1}{200}$)}};
\end{tikzpicture}
\endgroup
  \caption{Per-step weighting (Eq.~\ref{eq:pooled-occupancy}) versus per-trajectory weighting (Eq.~\ref{eq:two-occupancies}) for two equally likely task types whose student trajectories have $T(\tau_1)=2$ and $T(\tau_2)=100$ steps. Per-step weighting gives all $102$ steps the same mass, so type~2 supplies about $98\%$ of the teacher's starting contexts. Per-trajectory weighting gives each trajectory mass $\tfrac12$, split evenly over its steps, so both task types keep their weight $\mu$.}
  \label{fig:state-averaging}
\end{figure}

\subsection{Per-trajectory switching with fewer rollout playbacks}
\label{app:lazy-switch}
Algorithm~\ref{alg:sft-rollout} draws the teacher takeover time $t'\sim \mathrm{Unif}\{1,\ldots,T(\tau)\}$, which requires that each student rollout $\tau$ has completed. This necessitates either per-step rollout snapshotting or a \emph{post-hoc} playback of the environment state $s_{t'-1}$ and the LLM context $h_{t'-1}$, so that the teacher continues from the student's context exactly. Either case is costly and adds latency, so in this section we discuss how to implement our rule $\Tswitch(t\mid\tau)=1/T(\tau)$ to reduce the frequency of replaying the rollout.

Let $T_0$ be any upper bound on the student's rollout length, such as $\Tmax$, and replace the student rollout, the switch-time draw, and the snapshot playback (lines 7, 8, and 12 of Algorithm~\ref{alg:sft-rollout}) with the following.
\begin{enumerate}[leftmargin=1.8em,itemsep=1pt,topsep=2pt]
\item Before the rollout starts, draw a tentative switch time $\tilde t\sim\mathrm{Unif}\{1,\ldots,T_0\}$.
\item \emph{Live switch.} Roll out the student. If it is still running when step $\tilde t$ is due, i.e., $T(\tau)\ge\tilde t$, set $t'=\tilde t$ and let $\pie$ take over in the live environment. Nothing is restored, and the student's steps from $\tilde t$ onward are never generated.
\item \emph{Fallback.} If the student terminates first, i.e., $T(\tau)<\tilde t$, draw $t'\sim\mathrm{Unif}\{1,\ldots,T(\tau)\}$ and restore $(h_{t'-1},s_{t'-1})$ as in Algorithm~\ref{alg:sft-rollout}.
\end{enumerate}
Items~1 and 2 are OEC's sampling protocol (\S\ref{app:pooled-switch-equivalence}). Our addition is the fallback, where a $t'$ draw beyond the stopping time of $\tau$ is redrawn with guaranteed acceptance rather than discarded.

\paragraph{Equivalence.}
Fix a student rollout $\tau$ and a step $t\le T(\tau)$. The switch lands on $t$ either through the live switch ($\tilde t=t$) or through the fallback ($\tilde t>T(\tau)$, then $t'=t$), so
\begin{equation}
\Pr(t'=t\mid\tau)
=\underbrace{\frac{1}{T_0}}_{\text{live switch}}
+\underbrace{\frac{T_0-T(\tau)}{T_0}\cdot\frac{1}{T(\tau)}}_{\text{fallback}}
=\frac{1}{T(\tau)}.
\label{eq:lazy-switch}
\end{equation}
This is the weight $w_t(\tau)=1/T(\tau)$ of \S\ref{app:pooled-switch-equivalence}. In a live switch the rest of $\tau$ is never observed, but $\tilde t$ is drawn independently of the rollout, so the prefix has the same distribution as the prefix of a completed rollout, and the teacher's continuation depends on $\tau$ only through $(h_{t'-1},s_{t'-1})$, which the live environment provides just as a replayed snapshot would. The starting context and the continuation $\tau'$ therefore have the same joint distribution as under Algorithm~\ref{alg:sft-rollout}.

\paragraph{Savings.}
The live switch is taken with probability $\Pr(\tilde t\le T(\tau))=\mathbb E_{\tau}[T(\tau)]/T_0$. Per proposal, the procedure changes the expected number of playbacks and of student steps as follows:
\begin{align}
\text{playbacks:}\quad & 1\;\longrightarrow\;1-\frac{\mathbb E_{\tau}[T(\tau)]}{T_0},
\notag\\
\text{student steps:}\quad & \mathbb E_{\tau}[T(\tau)]\;\longrightarrow\;\mathbb E_{\tau}[T(\tau)]-\frac{\mathbb E_{\tau}\bigl[T(\tau)\,(T(\tau)+1)\bigr]}{2T_0}.
\label{eq:lazy-savings}
\end{align}
The second expression counts $\tilde t-1$ student steps in a live switch and $T(\tau)$ in the fallback. If every rollout uses the full budget $T_0$, no proposal needs a restore and about half of the student steps are saved. If rollouts are much shorter than $T_0$, almost every proposal falls back and the saving vanishes, so $T_0$ should be the tightest bound available.

The above procedure is applicable whenever the pre-filter $\PhiPre$ does not require to know how $\tau$ ends, which covers \OPshort{}, \OPfull{}, and \OPsuccess{}. \OPcritical{} filters on the student's terminal outcome, so it still needs to see the completed rollout.

\clearpage
\section{Experiment Details}
\label{app:exp-details}
\subsection{Compute Resources}
\textbf{HotpotQA.} All SFT and GRPO experiments were run on a single 8xH100 node with 64 CPUs. The \href{https://github.com/PeterGriffinJin/Search-R1/blob/main/docs/retriever.md}{local database} hosting the Wikipedia corpus was co-located with the \texttt{verl} trainer pipeline. The Qwen3-235B-A22B teacher model was queried via API calls through OpenRouter.

\textbf{ALFWorld.} SFT and GRPO experiments were run with one to two H100 GPUs and 32-64 CPUs.

\textbf{Terminal-Bench-Dev.} Most of Terminal-Bench-Dev SFT tasks were run on a single A100 node with 4 GPUs, and the rest on a single H100 node with 4 GPUs. The GLM 4.6 teacher model was queried via API calls through OpenRouter. Virtual CPU sandboxes were accessed through Daytona.

\subsection{HotpotQA}
\subsubsection{LLM Agent Inference Setup}\label{app:HotpotQA-data-setup}
\textbf{Prompts and interaction format.} The Qwen2.5-3B-Instruct student~\citep{qwen2025qwen25technicalreport} and Qwen3-235B-A22B teacher~\citep{yang2025qwen3technicalreport} use the \emph{same} system prompt and tool interface. The boxes below reproduce the system prompt, initial user message, search-response template, and invalid-action feedback. Only the placeholders highlighted in yellow are replaced for each instance, using the question and retrieved document titles and text. Each episode allows at most five assistant turns, including the final answer. This implies that to succeed, the student could query the database for at most four times. Teacher continuations inherit the student prefix and its remaining turn budget.

\textbf{Recovery from invalid actions.} In the original \texttt{verl} v0.4.0 + SGLang agent loop, an ill-formatted tool call or unclosed answer boxing (\texttt{<answer>...</answer>}) would kill the episode even when the agent has unused turn budget. We modified the loop to return an \emph{Invalid-action feedback} message instead. This feedback is enclosed in \texttt{<tool\_response>...</tool\_response>} tags, so the agent can act again if turn budget remain. Invalid actions still consume an assistant turn. The teacher's API loop uses the same feedback. This permits recovery from interface errors, as in a coding-agent interaction where a failed tool invocation returns an error that the agent can address in the next turn.

\textbf{Inference settings.} Teacher demonstrations and continuations used temperature $0.7$ and top-$p$ $0.95$. Student rollouts, development-set (``dev set'') evaluations used for SFT checkpoint selection, and final test evaluations all use temperature $1.0$, top-$p$ $1.0$, and a maximum response length of 3000 tokens counting the cumulative agent turns and tool responses. We believe this token limit is generously set, as in the final heldout rollouts that supplied the main text's performance metrics, we found that for every evaluated checkpoint, less than $1\%$ of rollouts exceeded this response token limit.

\begin{taskbox}[tbprompt, title after break={System prompt (continued)}]{System prompt}
\begin{trajverb}
You are Qwen, created by Alibaba Cloud. You are a helpful assistant.

## Instructions
- Answer the knowledge question given in the initial user message. After reasoning, if you find you lack some knowledge, you can call the search engine by writing a single JSON tool call inside <tool_call></tool_call>. The user will return the top searched results between <tool_response> and </tool_response>.
- Maximum search tool calls allowed: 4
- After at most 4 searches or as soon as you feel confident about the answer, provide your answer submission immediately in <answer> </answer> tags.
- Keep final answer concise - just the answer inside the XML tags, no explanation needed.
- Temporal context: Assume the current year is 2018


## 2 EXAMPLE FLOWS ##
### Example Flow A: Sufficient information
User:
Who was the first person to walk on the moon?

Assistant:
The user is asking about the first moon landing. I believe this was Neil Armstrong during Apollo 11, but let me verify.
<tool_call> {"name": "search", "arguments": {"query": "first person walk moon Apollo 11"}} </tool_call>

User:
[...tool_response confirms Neil Armstrong walked on the moon July 20, 1969...]

Assistant:
The search confirms the answer. No further queries needed.
<answer> Neil Armstrong </answer>

### Example Flow B: Insufficient information
User:
What year did the director of Inception appear in Time's 100 most influential people?

Assistant:
I need to find who directed Inception and when they appeared in Time's.
<tool_call> {"name": "search", "arguments": {"query": "Inception movie director"}} </tool_call>

User:
[...tool_response mentions Christopher Nolan directed Inception, but no mention of Times magazine...]

Assistant:
The search indicates Christopher Nolan directed Inception, but I still need his Times appearance.
<tool_call> {"name": "search", "arguments": {"query": "Christopher Nolan Times 100 influential people"}} </tool_call>

User:
[...tool_response suggests that Nolan was named one of Time magazine's 100 most influential people in 2015...]

Assistant:
Now I have the complete answer.
<answer> 2015 </answer>
## END OF 2 EXAMPLE FLOWS ##


# Tools

You may call one function to assist with the user query.

You are provided with function signatures within <tools></tools> XML tags:

<tools>
{"type": "function", "function": {"name": "search", "description": "Searches a Wikipedia database for relevant information based on the query you provide. The tool will return the three most semantically relevant document chunks for your query.", "parameters": {"type": "object", "properties": {"query": {"type": "string", "description": "Your query in concise natural language. Can be a phrase or a single word.", "enum": null}}, "required": ["query"]}, "strict": false}}
</tools>

For each function call, return a json object with function name and arguments within <tool_call></tool_call> XML tags:
<tool_call>
{"name": <function-name>, "arguments": <args-json-object>}
</tool_call>
\end{trajverb}
\end{taskbox}

\begingroup
\newcommand{\hpfield}[1]{{\setlength{\fboxsep}{1pt}\colorbox{yellow!35}{#1}}}
\begin{taskbox}[tbprompt, unbreakable]{First user turn (specifies the QA task)}
\begin{trajverb}[commandchars=+\[\], breakanywhere=false]
Question: +hpfield[{question}]
\end{trajverb}
\end{taskbox}

\begin{taskbox}[tbenv]{Search-response template}
\begin{trajverb}[commandchars=+\[\], breakanywhere=false]
<tool_response>
{"result": "Doc 1 (Title: +hpfield[{title_1}])\n+hpfield[{text_1}]\n\nDoc 2 (Title: +hpfield[{title_2}])\n+hpfield[{text_2}]\n\nDoc 3 (Title: +hpfield[{title_3}])\n+hpfield[{text_3}]"}
</tool_response>
\end{trajverb}
\end{taskbox}
\endgroup

\begin{taskbox}[tbenv]{Invalid-action feedback}
\begin{trajverb}
You produced an invalid action. Recall the system message's instruction for you. After planning your next step, you should EITHER call the search engine (enclosed in <tool_call> and </tool_call> tags with the correct schema) OR provide the final answer (enclosed in <answer> and </answer> tags).
\end{trajverb}
\end{taskbox}

\noindent In the boxes above, the fields highlighted in {\setlength{\fboxsep}{1pt}\colorbox{yellow!35}{yellow}} are placeholders. They are replaced by the actual content of each instance during inference.

\subsubsection{Model Training}
\label{app:HotpotQA-train-detail}

\begin{figure}[htbp]
\centering
\includegraphics[width=0.9\linewidth]{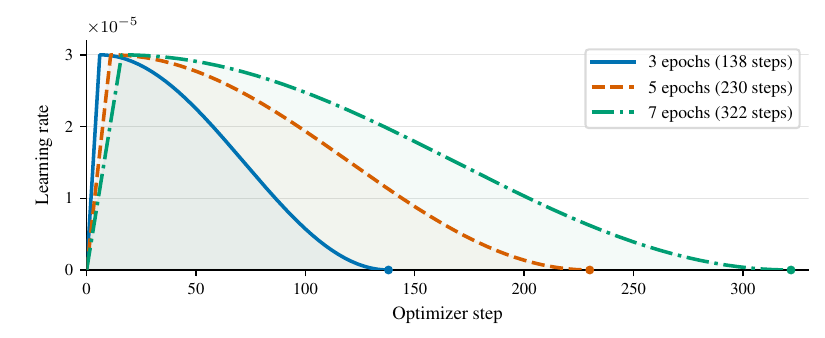}
\caption{HotpotQA \pureBC{} ($n=1$) learning-rate schedules for three training runs, holding the SFT training data seen in each epoch constant. Every training run uses 46 optimizer steps per epoch, with 5\% linear warmup followed by cosine decay.}
\label{fig:hotpotqa-lr-schedules}
\end{figure}

All HotpotQA experiments are trained on a single node with 8$\times$ H100 GPUs via the \texttt{verl} framework~\citep{verl} with FSDP2 sharding in BF16 mixed precision. Full-parameter finetuning is used.
Table~\ref{tab:HotpotQA_sft_hp} lists the SFT hyperparameters we consistently applied to all training runs. We sweep total training lengths of 3, 5, and 7 epochs to reduce sensitivity to the learning-rate schedule. Figure~\ref{fig:hotpotqa-lr-schedules} shows the corresponding cosine schedules for \pureBC{} ($n=1$, i.e. one teacher demo per task). Longer schedules have a larger area under the learning-rate curve, implying more weight updates. We save a model checkpoint after each epoch, and evaluate success rate on dev-set tasks to select the best checkpoint across a sweep, described more in \S\ref{app:HotpotQA-checkpoint-selection}.

\begin{table}[htbp]
\centering
\caption{HotpotQA SFT hyperparameters.}
\label{tab:HotpotQA_sft_hp}
\begin{tabular}{ll}
\toprule
\textbf{Hyperparameter} & \textbf{Value} \\
\midrule
Optimizer              & AdamW \\
Learning rate          & $3 \times 10^{-5}$ \\
LR scheduler           & Cosine \\
Warmup ratio           & 0.05 (i.e. 5\%) \\
Weight decay           & 0.01 \\
Optimizer betas        & $(0.9, 0.95)$ \\
Train batch size       & 64 \\
Micro batch size / GPU & 2 \\
Max sequence length    & 4096 \\
Number of epochs       & sweep: \{3, 5, 7\} \\
Precision              & BF16 \\
\bottomrule
\end{tabular}
\end{table}

\subsubsection{SFT Checkpoint Selection}
\label{app:HotpotQA-checkpoint-selection}
We select SFT checkpoints by task success under the student's own rollouts on held-out dev questions, evaluated after every epoch, and not by validation loss or early stopping. This choice follows our empirical observations and prior literature on LLM finetuning. \citet{zhou2023lima} find that held-out perplexity rises during finetuning while generation quality keeps improving, and therefore select checkpoints on a small development set. \citet{xia2024less} likewise report, across several models and tasks, that a lower validation loss does not always correspond to a higher task accuracy.

\begin{figure}[htbp]
\centering
\includegraphics[width=\linewidth]{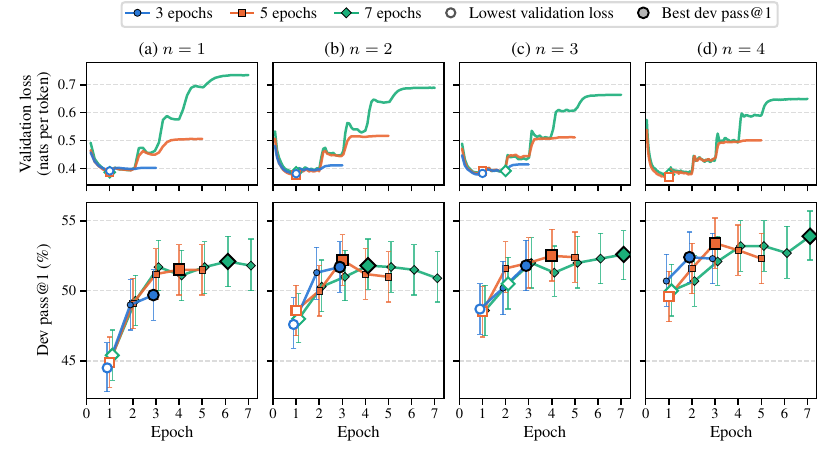}
\caption{We find that validation loss is a misleading checkpoint-selection signal for agent finetuning. (a)-(d) vary the number of distinct \pureBC{} teacher demos per HotpotQA task. \emph{Top}: mean NLL on held-out teacher demos, logged every 8 optimizer steps. \emph{Bottom}: dev pass@1 of each per-epoch checkpoint with 95\% confidence intervals (Table~\ref{app:hotpotqa-pure-bc-train-detail}). Hollow markers are the checkpoints with the lowest validation loss, and large outlined markers are those with the best dev pass@1. The validation loss rises in stairs at epoch boundaries, also observed by~\citet{hfforum2022stairs} and \citet{alpaca2023stairs}.}
\label{fig:hotpotqa-val-loss-vs-dev-acc}
\end{figure}

Our runs show the same pattern, which Figure~\ref{fig:hotpotqa-val-loss-vs-dev-acc} illustrates for the \pureBC{} baselines. The validation loss is the token-mean negative log-likelihood (NLL) on teacher demos for 818 held-out tasks that have no overlap with the train/dev/test set. The loss reaches its minimum after about one epoch and then rises with every further epoch, the usual signature of overfitting, while dev pass@1 keeps improving for several more epochs. In 10 of the 11 \pureBC{} runs with a logged validation curve, the first-epoch checkpoint has the lowest validation loss, whereas the best dev checkpoint always comes from epoch 3 or later. Selecting on validation loss, or stopping once it rises, would have lowered the dev pass@1 of these baselines by 2.1 to 6.7 points (median 3.9). The per-epoch dev sweep thus guards against underestimating \pureBC{}, and we apply the same procedure to every recipe.

Tables~\ref{app:hotpotqa-pure-bc-train-detail}--\ref{app:hotpotqa-filtered-train-detail} report the per-epoch checkpoint sweeps used for model selection. We compare pass@1 on 3,000 held-out medium-level questions, matching the difficulty of the training questions. 
Each dev evaluation uses one sample per question, so neighboring checkpoints often differ by less than their confidence intervals. How to select checkpoints reliably under a small evaluation budget remains an open experiment design question.

\begingroup
\setlength{\intextsep}{6pt plus 2pt minus 2pt}
\newcommand{\HPTableStyle}{%
  \centering
  \setlength{\tabcolsep}{3.5pt}%
  \renewcommand{\arraystretch}{1.0}%
  \fontsize{9}{10.5}\selectfont
}
\makeatletter
\let\HPInput\@@input
\let\CT@do@color\relax
\makeatother

\newcommand{\HPScore}[3]{\shortstack[c]{#1\\[-1.3pt]{\fontsize{7}{8}\selectfont\color{black!65}#2--#3}}}
\newcommand{\HPBest}[3]{\HPScore{\textbf{#1}}{#2}{#3}}
\newsavebox{\HPSelectedBox}
\newcommand{\HPSelected}[3]{{%
  \setlength{\fboxsep}{0.8pt}%
  \setlength{\fboxrule}{0.35pt}%
  \sbox{\HPSelectedBox}{\HPBest{#1}{#2}{#3}}%
  \makebox[\wd\HPSelectedBox][c]{%
    \raisebox{0pt}[\ht\HPSelectedBox][\dp\HPSelectedBox]{\fbox{\usebox{\HPSelectedBox}}}}%
}}
\newcommand{\HPNA}{}
\newsavebox{\HPScoreBox}
\newcommand{\HPFail}{%
  \sbox{\HPScoreBox}{\HPScore{00.0}{00.0}{00.0}}%
  \raisebox{\dimexpr(\ht\HPScoreBox-\dp\HPScoreBox-\height+\depth)/2\relax}{\textcolor{black!65}{\textit{Fail}}}%
}
\newcommand{\HPSweep}[4]{#1 & #2 & #3 & \textbf{#4}\\[0pt]}
\newcommand{\HPRecipeBreak}{\arrayrulecolor{black!30}\specialrule{0.25pt}{0.5pt}{4pt}\arrayrulecolor{black}}
\newcommand{\HPGroup}[1]{\addlinespace[3pt]\multicolumn{10}{@{}l}{\textbf{#1}}\\*[3pt]}
\newcommand{\HPHeader}[1]{%
  #1 & \shortstack{Total\\epochs} &
  \multicolumn{7}{c}{Checkpoint epoch} & Peak\\
  \cmidrule(lr){3-9}
  & & 1 & 2 & 3 & 4 & 5 & 6 & 7 & \\
}

\begin{table}[!htbp]
\HPTableStyle
\caption{HotpotQA SFT checkpoint selection for \pureBC{} on 3,000 medium-level dev questions. Total epochs specifies each run's cosine learning-rate schedule. Entries are pass@1 (\%) with 95\% confidence intervals beneath. \textbf{Bold} marks each run's best checkpoints. \boxed{Boxes} identify checkpoints selected for final \texttt{test} split evaluation and reporting in \S\ref{sec:main-results}. Peak is the highest score in each run.}
\label{app:hotpotqa-pure-bc-train-detail}
\begin{tabular}{@{}lc*{7}{c}c@{}}
\toprule
\HPHeader{BC data}
\midrule
\HPGroup{\pureBC{} (full teacher demos)}
\HPInput{tables/hotpotqa_checkpoints/pure_bc.tex}
\bottomrule
\end{tabular}
\par\smallskip
\begin{minipage}{\textwidth}
\fontsize{8}{9.2}\selectfont
\textit{Notes.} BC data denote unique questions $\times$ i.i.d. teacher trajectories per question. Blank cells are beyond the training schedule. \textit{Fail} denotes a crashed evaluation.
\end{minipage}
\end{table}

\begin{table}[!htbp]
\HPTableStyle
\caption{HotpotQA SFT checkpoint selection for \OPshort{}, varying the teacher inference cap $\Ki$ (\S\ref{sec:methods-data}). Evaluation setup, training schedules, and reporting conventions follow Table~\ref{app:hotpotqa-pure-bc-train-detail}.}
\label{app:hotpotqa-train-detail}
\begin{tabular}{@{}lc*{7}{c}c@{}}
\toprule
\HPHeader{\(\Ki\)}
\midrule
\HPInput{tables/hotpotqa_checkpoints/op_short.tex}
\bottomrule
\end{tabular}
\end{table}

\clearpage
\begin{table}[!htbp]
\HPTableStyle
\caption{HotpotQA SFT checkpoint selection for \OPsuccess{} and \OPcritical{}, varying the cap $\Ktr$ on teacher turns retained for SFT (\S\ref{sec:methods-data}). Evaluation setup, training schedules, and reporting conventions follow Table~\ref{app:hotpotqa-pure-bc-train-detail}.}
\label{app:hotpotqa-filtered-train-detail}
\begin{tabular}{@{}lc*{7}{c}c@{}}
\toprule
\HPHeader{\(\Ktr\)}
\midrule
\HPGroup{\OPsuccess{}}
\HPInput{tables/hotpotqa_checkpoints/op_success.tex}
\HPGroup{\OPcritical{}}
\HPInput{tables/hotpotqa_checkpoints/op_critical.tex}
\bottomrule
\end{tabular}
\end{table}
\endgroup

\clearpage

\subsection{ALFWorld}
\subsubsection{LLM Agent Inference Setup}\label{app:ALFWorld-data-setup}
\textbf{Prompts and interaction format.} The Qwen2.5-0.5B-Instruct student and the Qwen3-8B teacher receive the same user message template at every turn, which we adapt from \texttt{verl-agent}~\citep{verl-agent}. The initial context $q$ the agent sees is the user-turn template filled with the initial observation and the initial admissible actions. This functions like a system prompt. Every later user turn fills the same template with the new observation returned by the environment (e.g., ``You pick up the soapbottle 1 from the countertop 1.'') and the current list of admissible actions. Earlier turns are not summarized and stay in the context as chat history. The agent replies with its reasoning in \texttt{<think>} tags, followed by one action in \texttt{<action>} tags. The text between the action tags, lowercased and stripped of surrounding whitespace, is sent to the environment. Each episode allows at most $\Tmax=50$ agent steps. Teacher continuations inherit the student prefix and its remaining step budget.

\textbf{Invalid actions.} A response is invalid in format if it lacks a pair of \texttt{<action>} tags or contains Chinese characters, a rule we inherited from \texttt{verl-agent}. The environment answers every command it cannot execute with a null observation ``\texttt{Nothing happens.}'', which is returned in the usual user-turn template. This includes well-formatted actions outside the admissible list. The episode continues and the turn counts towards the turn limit, so the agent can recover in the next turn. In the final heldout rollouts that supplied the main text's performance metrics, $0.5\%$ of assistant turns were invalid in format and $11.8\%$ were well-formed but not admissible.

\textbf{Inference settings.} Teacher demonstrations, student rollouts, teacher continuations, dev-set evaluations used for SFT checkpoint selection, and final test evaluations all use temperature $1.0$ and top-$p$ $0.95$, with at most 2{,}048 new tokens per assistant turn and a context window of 25{,}600 tokens. In the final heldout rollouts, $1.9\%$ of episodes ended unsuccessfully by exhausting this context window before the turn limit (between $1.0\%$ and $3.5\%$ per evaluated checkpoint). The teacher runs in Qwen3's thinking mode, so its \texttt{<think>} blocks are part of the demonstrations. For the Qwen2.5 student, these tags are ordinary text.

\begingroup
\newcommand{\alffield}[1]{{\setlength{\fboxsep}{1pt}\colorbox{yellow!35}{#1}}}
\begin{taskbox}[tbprompt, unbreakable]{User turn (every turn)}
\begin{trajverb}[commandchars=+\(\), breakanywhere=false]
Your current observation is: +alffield({observation})
Your admissible actions of the current situation are: [+alffield({admissible_actions})].

Now it's your turn to take an action.
You should first reason step-by-step about the current situation. This reasoning process MUST be enclosed within <think> </think> tags.
Once you've finished your reasoning, you should choose an admissible action for current step and present it within <action> </action> tags.
\end{trajverb}
\end{taskbox}

\begin{taskbox}[tbenv]{Initial observation (specifies the task)}
\begin{trajverb}[commandchars=+\(\), breakanywhere=false]
-= Welcome to TextWorld, ALFRED! =-

You are in the middle of a room. Looking quickly around you, you see +alffield({receptacles}).

Your task is to: +alffield({task})
\end{trajverb}
\end{taskbox}

\begin{taskbox}[tbenv]{Admissible-action list format}
\begin{trajverb}[commandchars=+\(\), breakanywhere=false]
'+alffield({action_1})'
 '+alffield({action_2})'
 ...
 '+alffield({action_n})'
\end{trajverb}
\end{taskbox}

\begin{taskbox}[colframe=black!55, colback=black!3, boxed title style={colback=black!55}, unbreakable]{Response format (teacher and student)}
\begin{trajverb}[commandchars=+\(\), breakanywhere=false]
<think>
+alffield({reasoning})
</think>

<action>+alffield({action})</action>
\end{trajverb}
\end{taskbox}
\endgroup

\noindent In the boxes above, the fields highlighted in {\setlength{\fboxsep}{1pt}\colorbox{yellow!35}{yellow}} are placeholders. They are replaced by the actual content of each instance during inference.

\subsubsection{Model training}
\label{app:alf-train-detail}
ALFWorld experiments are also built upon \texttt{VeRL} repo.
We detail the experiment configs for SFT in Table~\ref{tab:alf_sft_hp}.
The maximum sequence length of 26{,}112 is slightly above the 25{,}600-token context window used at inference (\S\ref{app:ALFWorld-data-setup}); the extra 512 tokens absorb the overhead of re-tokenizing teacher outputs, generated with the Qwen3 tokenizer, back into the student's Qwen2.5 tokenizer for SFT.

\begin{table}[htbp]
\centering
\caption{ALFWorld SFT hyperparameters.}
\label{tab:alf_sft_hp}
\begin{tabular}{ll}
\toprule
\textbf{Hyperparameter} & \textbf{Value} \\
\midrule
Optimizer              & AdamW \\
Learning rate          & $5 \times 10^{-5}$ \\
LR scheduler           & Cosine \\
Warmup ratio           & 0.1 (i.e. 10\%) \\
Weight decay           & 0.01 \\
Optimizer betas        & $(0.9, 0.95)$ \\
Train batch size       & 16 \\
Micro batch size / GPU & 2 \\
Max sequence length    & 26112 \\
Number of epochs       & 8 \\
Precision              & BF16 \\
\bottomrule
\end{tabular}
\end{table}

\subsubsection{SFT Checkpoint Selection}
\label{app:ALFWorld-checkpoint-selection}
Each of the 15 ALFWorld recipes in \S\ref{sec:main-results} is trained once with the 8-epoch cosine schedule of Table~\ref{tab:alf_sft_hp}, so unlike HotpotQA (\S\ref{app:HotpotQA-checkpoint-selection}) there is no sweep over schedule lengths. We save 16 checkpoints per run, one every half epoch, and evaluate each on a dev set of 96 games from the ALFWorld \texttt{train} split, 16 per task type and disjoint from the 32 training games. Every checkpoint plays 8 rollouts per game ($96 \times 8 = 768$ episodes) at temperature $1.0$ and top-$p$ $0.95$. The checkpoint with the highest dev success rate is then evaluated on the test set (\texttt{valid\_unseen}, 134 games), which has no game overlap with the dev set.

\begin{figure}[htbp]
\centering
\includegraphics[width=\linewidth]{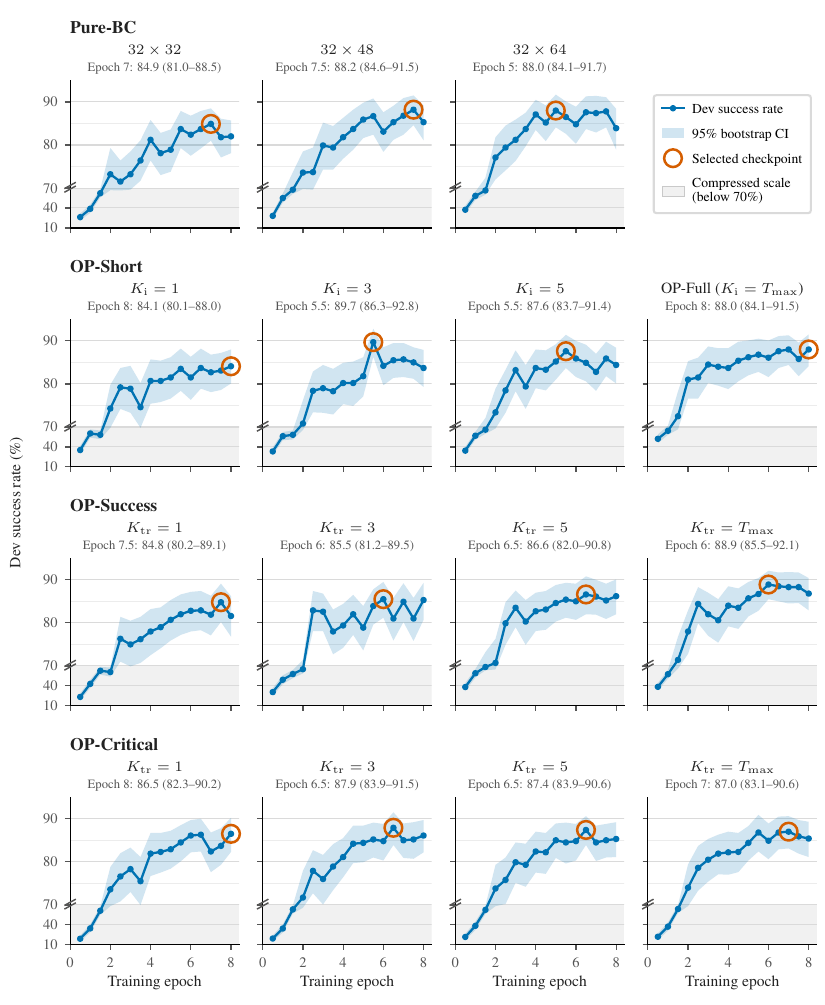}
\caption{ALFWorld SFT checkpoint selection on 96 dev games (8 rollouts per game to reduce stochasticity). Each panel is one SFT run with 16 checkpoints; rows group runs by data recipe. \pureBC{} panels are labeled by tasks $\times$ teacher demonstrations per task, \OPshort{} panels by the teacher inference cap $\Ki$, and \OPsuccess{} and \OPcritical{} panels by the retention cap $\Ktr$ (\S\ref{sec:methods-data}). Lines show dev success rate (\%) and bands show 95\% bootstrap confidence intervals. The vertical axis is compressed below 70\% (gray region) to enlarge the range where the later checkpoints lie. Circles mark the checkpoints selected for final test evaluation and reporting in \S\ref{sec:main-results}; the line under each panel title gives the selected epoch, its success rate, and its interval.}
\label{fig:alfworld-dev-sweeps}
\end{figure}

Figure~\ref{fig:alfworld-dev-sweeps} reports all 240 checkpoint evaluations, one panel per run, and circles the selected checkpoints. The shaded bands are task-level 95\% bootstrap intervals: we average the 8 rollouts of each game, resample the 96 game means with replacement 10{,}000 times, and take the 2.5th and 97.5th percentiles. All selected checkpoints lie between epochs 5 and 8.

\clearpage

\subsection{OpenThoughts-TB-Dev}
\label{tb-train-detail}
\subsubsection{LLM Agent Inference Setup}\label{app:TBDev-data-setup}\nobreak
\textbf{Prompts and interaction format.} We evaluate on TB-Dev with the \texttt{Terminus-2} agent harness~\citep{merrill2026terminal}, as implemented in the \href{https://github.com/harbor-framework/harbor}{Harbor} framework and run in its JSON-response mode. The first message the LLM agent sees is a user turn that fills the template in the first box below with the task description and the initial terminal screen. This message specifies the response format and the starting state, so it functions like a system prompt. The LLM agent replies with one JSON object containing an analysis, a plan, and a batch of keystroke commands. The harness types the keystrokes into a \texttt{tmux} session inside the task's sandbox, waits the requested time, and returns the terminal text produced since the previous step as the next user turn. Training data collection uses the same prompt format, and Appendix~\ref{app:op-example} shows an \OPshort{} training trajectory.

\textbf{Invalid responses and task completion.} A failing shell command triggers no special message, since its error text is part of the terminal output. The harness gives feedback only on the format of a reply. If a reply contains no parsable JSON object or lacks a required field, no command is executed and the next user turn is the parse-error feedback shown below. Replies with minor defects, such as extra text around the JSON object, are still executed, with a warning prepended to the terminal output. When the agent first sets \texttt{task\_complete} to true, the harness asks for confirmation and ends the episode only if the next reply sets it again. Evaluation imposes no turn limit, so an episode otherwise ends when the task's time budget runs out. The task's verifier then runs its tests on the final sandbox state.

\textbf{Inference settings.} Evaluation samples at temperature $0.7$, the Terminus-2 harness, with top-$p$ $0.95$ and top-$k$ $20$ from the Qwen3 generation config. The agent works within a context limit of 32{,}768 tokens. Each Daytona sandbox has 1 CPU and 2 GB of memory. For the final heldout evaluation, every checkpoint is evaluated with three independently sampled rollouts per task to reduce randomness and estimate standard error.

\begingroup
\newcommand{\tbdfield}[1]{{\setlength{\fboxsep}{1pt}\colorbox{yellow!35}{#1}}}
\begin{taskbox}[tbprompt, title after break={First user turn (continued)}]{First user turn (specifies the response format and the task)}
\begin{trajverb}[commandchars=|\@\!, breakanywhere=false]
You are an AI assistant tasked with solving command-line tasks in a Linux environment. You will be given a task description and the output from previously executed commands. Your goal is to solve the task by providing batches of shell commands.

Format your response as JSON with the following structure:

{
  "analysis": "Analyze the current state based on the terminal output provided. What do you see? What has been accomplished? What still needs to be done?",
  "plan": "Describe your plan for the next steps. What commands will you run and why? Be specific about what you expect each command to accomplish.",
  "commands": [
    {
      "keystrokes": "ls -la\n",
      "duration": 0.1
    },
    {
      "keystrokes": "cd project\n",
      "duration": 0.1
    }
  ],
  "task_complete": true
}

Required fields:
- "analysis": Your analysis of the current situation
- "plan": Your plan for the next steps
- "commands": Array of command objects to execute

Optional fields:
- "task_complete": Boolean indicating if the task is complete (defaults to false if not present)

Command object structure:
- "keystrokes": String containing the exact keystrokes to send to the terminal (required)
- "duration": Number of seconds to wait for the command to complete before the next command will be executed (defaults to 1.0 if not present)

IMPORTANT: The text inside "keystrokes" will be used completely verbatim as keystrokes. Write commands exactly as you want them sent to the terminal:
- You must end every command with a newline (\n) or it will not execute.
- For special key sequences, use tmux-style escape sequences:
  - C-c for Ctrl+C
  - C-d for Ctrl+D

The "duration" attribute specifies the number of seconds to wait for the command to complete (default: 1.0) before the next command will be executed. On immediate tasks (e.g., cd, ls, echo, cat) set a duration of 0.1 seconds. On commands (e.g., gcc, find, rustc) set a duration of 1.0 seconds. On slow commands (e.g., make, python3 [long running script], wget [file]) set an appropriate duration as you determine necessary.

It is better to set a smaller duration than a longer duration. It is always possible to wait again if the prior output has not finished, by running {"keystrokes": "", "duration": 10.0} on subsequent requests to wait longer. Never wait longer than 60 seconds; prefer to poll to see intermediate result status.

Important notes:
- Each command's keystrokes are sent exactly as written to the terminal
- Do not include extra whitespace before or after the keystrokes unless it's part of the intended command
- Extra text before or after the JSON will generate warnings but be tolerated
- The JSON must be valid - use proper escaping for quotes and special characters within strings
- Commands array can be empty if you want to wait without taking action

Task Description:
|tbdfield@{instruction}!

Current terminal state:
Current Terminal Screen:
|tbdfield@{terminal_screen}!
\end{trajverb}
\end{taskbox}

\begin{taskbox}[tbenv]{Terminal output (every later user turn)}
\begin{trajverb}[commandchars=|\@\!, breakanywhere=false]
New Terminal Output:
|tbdfield@{new_terminal_text}!
\end{trajverb}
\end{taskbox}

\begin{taskbox}[tbenv]{Parse-error feedback (no command is executed)}
\begin{trajverb}[commandchars=|\@\!, breakanywhere=false]
Previous response had parsing errors:
ERROR: |tbdfield@{error}!

Please fix these issues and provide a proper JSON response.
\end{trajverb}
\end{taskbox}

\begin{taskbox}[tbenv]{Task-completion confirmation}
\begin{trajverb}[commandchars=|\@\!, breakanywhere=false]
Current terminal state:
|tbdfield@{terminal_output}!

Are you sure you want to mark the task as complete? This will trigger your solution to be graded and you won't be able to make any further corrections. If so, include "task_complete": true in your JSON response again.
\end{trajverb}
\end{taskbox}
\endgroup

\noindent In the boxes above, the fields highlighted in {\setlength{\fboxsep}{1pt}\colorbox{yellow!35}{yellow}} are placeholders. They are replaced by the actual content of each instance during inference. In the confirmation message, \texttt{\{terminal\_output\}} is a complete terminal-output message.

\Needspace{12\baselineskip}
\subsubsection{Model Training}\label{app:TBDev-train-detail}\nobreak
Table~\ref{tab:tbdev_sft_hyperparameters} describes the training pipeline of our OpenThought-TB-Dev experiment. The same hyperparameters are used for all the data recipes we test, i.e. \pureBC{} and \OPshort{} with varying $\Ki$ values. This training pipeline and configuration are inherited from the OT-Agent project's codebase, the former being a fork of the finetuning framework \texttt{Llama-Factory}.
We use full-parameter finetuning in BF16 mixed precision, with DeepSpeed ZeRO-3 sharding across 4 GPUs. Because the \texttt{v1-SFT} training tasks (\texttt{nl2bash} \& \texttt{inferredbugs}) do not include answer verifiers from which to build a rollout-based dev set, each run trains for a fixed 7 epochs and uses the final checkpoint directly without a checkpoint-selection step, unlike HotpotQA and ALFWorld. This is also OT-Agent project's official setup as of May 2026. TB-Dev (despite its name) is reserved strictly as the test set.

\begin{table}[ht]
\centering
\caption{OpenThoughts-TB-Dev SFT training configuration.}
\label{tab:tbdev_sft_hyperparameters}
\begin{tabular}{ll}
\toprule
\textbf{Hyperparameter} & \textbf{Value} \\
\midrule
Learning rate & $4 \times 10^{-5}$ \\
Train batch size per device & 1 \\
Gradient Accumulation & 4 \\
Seed & 42 \\
Distributed type & Multi-GPU \\
Number of devices & 4 \\
Total train batch size & 16 \\
Optimizer & AdamW Torch Fused \\
Optimizer betas & $(0.9, 0.98)$ \\
Optimizer epsilon & $1 \times 10^{-8}$ \\
Weight decay & 0 \\
Learning rate scheduler & Cosine \\
Warmup ratio & 0.1 \\
Max sequence length & 32768 \\
Number of epochs & 7 \\
Precision & BF16 \\
\bottomrule
\end{tabular}
\end{table}

\subsubsection{TB-Dev extra experiments}

Table \ref{tab:qwen3_8b_tbdev_sft} shows all TB-dev experiment results in different settings. First two pure OP results validates our conclusion that the optimal continuation length should be set with the task horizon and is not a fixed number. We would like to mention that our method is not exclusive to OEC \citep{lauffer2025oec}. Instead, one can regard our method as a better initialization point than pure BC in OEC. 

\begin{table}[ht]
\centering
\small
\caption{Comparison of TB-dev evaluation scores for baseline models and Qwen3-8B under different SFT and RL settings. (numbers with * are from \href{https://web.archive.org/web/20251219014649/https://www.open-thoughts.ai/blog/agent}{OpenThoughts-Agent}, others are from our implementation. TB-dev score is averaged across three independent pass@1 evaluations.)}
\label{tab:qwen3_8b_tbdev_sft}
\resizebox{\linewidth}{!}{
\begin{tabular}{cccccccc}
\toprule
\textbf{Base Model} 
& \textbf{Teacher Model} 
& \textbf{Type}
& \textbf{Prefix Training Data}
& \textbf{BC Data}
& \textbf{On-policy Data} 
& \textbf{RL} 
& \textbf{TB-dev Score ($\%, \uparrow$)} \\
\midrule
\multirow{13}{*}{Qwen3-8B}
& \multirow{7}{*}{GLM-4.6}
& \multirow{7}{*}{Pure $\BC$}
& $10\%$
& $10\%$
& \multirow{7}{*}{0}
& 0
& 11.4 \\

& 
& 
& $25\%$
& $25\%$
& 
& 0 
& 10.2 \\

& 
& 
& $50\%$
& $50\%$
& 
& 0 
& 11.1 \\

& 
& 
& $75\%$
& $75\%$
& 
& 0 
& 13.6 \\

& 
& 
& $100\%$
& $100\%$
& 
& 0 
& 13.2 \\

& 
& 
& $100\%$
& $100\%$
& 
& 760 tasks 
& 16.7 \\

& 
& 
& $100\%$
& $100\%$
& 
& 760 tasks
& $17.3^*$ \\

\cmidrule(lr){2-8}

& \multirow{6}{*}{GLM-4.6}
& Pure OP, $\Ki=3$
& 10\% 
& 0\%
& 10\%
& 0 
& 14.7 \\

& 
& Pure OP, $\Ki=\Tmax$
& 10\% 
& 0\%
& 10\%
& 0 
& 16.7 \\

& 
& $\OPshort$, $\Ki=1$
& 10\% 
& 5\%
& 5\%
& 0 
& 11.3 \\

& 
& $\OPshort$, $\Ki=3$
& 10\% 
& 2\%
& 2\%
& 0 
& 16.1 \\

& 
& $\OPfull$, $\Ki=\Tmax$
& 10\% 
& 2\%
& 2\%
& 0 
& 18.0 \\

& 
& $\OPshort$, $\Ki=3$
& 10\% 
& 5\%
& 5\%
& 0 
& 19.1 \\

\midrule
Qwen3-8B
& \multicolumn{6}{c}{\multirow{4}{*}{Baselines}}
& $5.7^*$ \\

Qwen3-235B-A22B
& \multicolumn{6}{c}{}
& $26.9^*$ \\

GLM-4.6
& \multicolumn{6}{c}{}
& $29.9$ \\

GPT-5 Nano
& \multicolumn{6}{c}{}
& $37.4^*$ \\
    
\bottomrule
\end{tabular}
}
\end{table}

\section{Ablation Study Details}
\label{app:ablation-details}

\subsection{Reverse-KL OPD on ALFWorld}
\label{app:ablation-opd}

\textbf{Setup.} The student is \texttt{Qwen3-0.6B} and the teacher $\pie$ is the RL-finetuned \texttt{Qwen3-8B} model of \S\ref{sec:lab-data}, both hosted locally on H100 GPUs. We sampled 128 training tasks from the \texttt{train} split and collected one teacher demonstration per task. The initial student $\pi_1$ is trained by SFT on a random 64 subset of these demonstrations with hyperparameters of Table~\ref{tab:alf_sft_hp}. The behavior policy of both OPD and \OPshort{} initializes from this same $\pi_1$, so its creation cost is excluded from the comparison.

\textbf{OPD.} We use the reverse-KL on-policy distillation \href{https://web.archive.org/web/20260720102941/https://verl.readthedocs.io/en/latest/algo/opd.html}{\texttt{verl} recipe}, which follows the policy-gradient formulation of MiniLLM \citep{gu2024minillm}. In each epoch, the current student $\pi_\theta$ rolls out once on each of the 128 tasks, and the teacher makes one forward pass on the resulting student trajectories to provide the teacher's next-token distribution on every student prefix (token-level). Each student token $a$ generated in context $h$ receives the advantage $\log \pie(a \mid h) - \log \pi_\theta(a \mid h)$, clamped to $[-10, 10]$, and no task outcome reward is used. Each batch of 16 trajectories receives a single gradient update, so the PPO-clip surrogate reduces to a plain policy gradient. We train for 12 epochs (96 optimizer steps) on three H100 GPUs, two for the student and one for the teacher, and save one checkpoint per epoch. Table~\ref{tab:alfworld_opd_hparams} lists the training hyperparameters.

\begin{table}[htbp]
\centering
\caption{Reverse-KL OPD hyperparameters on ALFWorld.}
\label{tab:alfworld_opd_hparams}
\begin{tabular}{ll}
\toprule
\textbf{Hyperparameter} & \textbf{Value} \\
\midrule
Distillation loss             & Reverse KL, policy-gradient estimator \\
Per-token advantage           & $\log \pie(a \mid h) - \log \pi_\theta(a \mid h)$, clamped to $[-10, 10]$ \\
Policy loss                   & PPO-clip surrogate, $\epsilon = 0.2$ \\
Gradient updates per batch    & 1 \\
Entropy coefficient           & 0 \\
Loss aggregation              & Token mean \\
\midrule
Optimizer                     & AdamW \\
Learning rate                 & $1 \times 10^{-6}$ \\
LR scheduler                  & Constant, no warmup \\
Optimizer betas               & $(0.9, 0.999)$ \\
Weight decay                  & 0.01 \\
Gradient clipping (max norm)  & 1.0 \\
Train batch size              & 16 trajectories \\
Micro batch size / GPU        & 2 \\
Number of epochs              & 12 (96 optimizer steps) \\
Precision                     & BF16 \\
\midrule
Student sampling              & Temperature 1.0, top-$p$ 0.95 \\
Max assistant turns per episode & 50 \\
Max context length            & $16{,}384$ \\
\bottomrule
\end{tabular}
\end{table}

\textbf{\OPshort{}.} The frozen $\pi_1$ rolls out three times on each task, which yields 384 student trajectories. We sample one switch time per trajectory, and the teacher continues from it for at most $\Ki=3$ turns. The base student $\pi_0$ is then trained by SFT for 8 epochs on a combination of the original 128 full teacher demonstrations and the 384 teacher continuations (Table~\ref{tab:alf_sft_hp}), and we report the final checkpoint. The 64 demonstrations that $\pi_1$ did not use are charged to \OPshort{}.

\textbf{Evaluation.} Every checkpoint is evaluated after training in a separate job on the \texttt{valid\_seen} (140 tasks) and \texttt{valid\_unseen} (134 tasks) test splits, with 4 rollouts per task at temperature 1 and top-$p$ $0.95$. Under this protocol the teacher reaches $90.4\%$ and $87.1\%$. Each method is trained once. Table~\ref{tab:ablation-opd-epochs} lists the scores plotted in Figure~\ref{fig:ablation-h100-epochs}.

\begin{table}[htbp]
\centering
\small
\caption{ALFWorld success rate (\%) after each training epoch. $\pi_1$ scores $69.8$ on \texttt{valid\_seen} and $71.5$ on \texttt{valid\_unseen}. The best score in each row is in bold.}
\label{tab:ablation-opd-epochs}
\setlength{\tabcolsep}{4pt}
\begin{tabular}{ll*{12}{c}}
\toprule
& Epoch & 1 & 2 & 3 & 4 & 5 & 6 & 7 & 8 & 9 & 10 & 11 & 12 \\
\midrule
\multirow{2}{*}{OPD}
& seen   & 71.4 & 69.3 & 67.1 & 64.3 & 63.6 & 63.2 & 65.0 & 68.0 & 68.8 & \textbf{73.0} & 71.6 & 71.3 \\
& unseen & 72.2 & 67.9 & 65.9 & 62.3 & 65.3 & 66.2 & 66.0 & 68.3 & 68.1 & 71.1 & \textbf{72.9} & 71.3 \\
\midrule
\multirow{2}{*}{\OPshort{}}
& seen   & 53.6 & 52.3 & 60.7 & 72.9 & 80.9 & 79.8 & 76.3 & \textbf{82.7} & & & & \\
& unseen & 58.4 & 56.9 & 57.7 & 71.6 & 75.6 & 76.9 & 73.1 & \textbf{77.2} & & & & \\
\bottomrule
\end{tabular}
\end{table}

\textbf{GPU time.} We measure cost in H100-hours, the wall-clock time of each job multiplied by its GPU count. Job startup, checkpoint saving, and teardown are excluded for both methods. Table~\ref{tab:ablation-opd-cost} itemizes the \OPshort{} pipeline, which totals $2{,}395$ GPU-seconds, or $0.67$ H100-hours. Counting the excluded overheads raises this to $0.77$ H100-hours. An OPD epoch takes 843 to $1{,}090$ seconds on three GPUs. The first epoch costs $0.70$ H100-hours and the 12 epochs together cost $9.9$ H100-hours, about $15\times$ the \OPshort{} total.

\begin{table}[htbp]
\centering
\small
\caption{Cost of the \OPshort{} ($\Ki=3$) pipeline in the OPD comparison. Tokens are all tokens processed by the model in that stage. Teacher continuations include $1.61$M tokens of student-prefix prefill. The collection job (rollouts and continuations) uses one H100 and the SFT job uses two for 594 seconds. The demonstration entry is the token-weighted share of the demonstration job for the 64 tasks outside $\pi_1$'s training set. FLOP entries are rounded independently.}
\label{tab:ablation-opd-cost}
\begin{tabular}{lccc}
\toprule
Stage & GPU-seconds & Tokens (M) & FLOPs ($10^{16}$) \\
\midrule
Student rollouts (384)                 & 540        & 3.02 & 0.4 \\
Teacher continuations                  & 311        & 2.06 & 3.1 \\
Teacher demonstrations (64 tasks)      & 356        & 0.43 & 0.7 \\
SFT (8 epochs)                         & $1{,}188$  & 23.2 & 8.3 \\
\midrule
Total                                  & $2{,}395$  &      & 12.4 \\
\bottomrule
\end{tabular}
\end{table}

\textbf{FLOPs.} As a hardware-independent check, we also estimate FLOPs as $2NT$ for inference and $6NT$ for training, where $T$ is the number of tokens processed and $N$ counts non-embedding parameters plus the output head ($0.60 \times 10^{9}$ for the student and $7.57 \times 10^{9}$ for the teacher). Teacher prefill tokens are counted. \OPshort{} totals $1.24 \times 10^{17}$ FLOPs, of which SFT accounts for $67\%$ and teacher inference for $30\%$ (Table~\ref{tab:ablation-opd-cost}). In OPD, each trajectory token is generated by the student, scored by the teacher, re-scored by the student, and then trained on, for $(10 N_{\mathrm{s}} + 2 N_{\mathrm{t}})$ FLOPs per token. With $0.96$M to $1.13$M trajectory tokens per epoch, an OPD epoch costs $2.0$ to $2.4 \times 10^{16}$ FLOPs. OPD's cumulative FLOPs pass the \OPshort{} total during epoch 6 and reach $2.66 \times 10^{17}$ after 12 epochs, $2.1\times$ the \OPshort{} total. Alternative conventions, such as counting embedding parameters or charging all 128 demonstrations to \OPshort{}, keep the crossing between epochs 5 and 7.

\textbf{Matched budgets.} In H100-hours, running the full \OPshort{} generate-data-and-train pipeline costs about as much as one OPD epoch. In FLOPs, it costs about as much as six. The gap reflects GPU utilization. OP-Short decouples data generation from training: student rollouts and teacher continuations are collected once, with many episodes served in parallel, and SFT then trains on the fixed dataset, so each stage keeps its GPUs busy. OPD instead couples the two. Every gradient step must wait for a fresh batch of rollouts from the current student, and while it waits, the teacher and the student trainer are mostly idle.

\subsection{RL on HotpotQA}
\label{app:ablation-rl}

This comparison asks how much RL compute the initial student $\pi_1$ needs to reach the accuracy that \OPshort{} attains with SFT alone. Both pipelines start from the same $\pi_1$, the \pureBC{} student trained on one teacher demonstration per training question (\S\ref{sec:lab-data}), so the cost of obtaining $\pi_1$ cancels. We express every cost in node-hours of a single $8\times$H100 node. Let $\pi_1^{(i)}$ denote $\pi_1$ after $i$ RL optimizer steps and $c_{\mathrm{RL}}(i)$ the cumulative RL training time through optimizer step $i$. We report
\begin{equation}
c_{\mathrm{RL}}(i^\star), \qquad i^\star = \min\bigl\{\, i \in \mathcal{I}_{\mathrm{eval}} : \mathrm{acc}\bigl(\pi_1^{(i)}\bigr) > \mathrm{acc}(\pi_2) \,\bigr\},
\label{eq:rl-match}
\end{equation}
where $\pi_2$ is the \OPshort{} ($\Ki=1$) student, $\mathcal{I}_{\mathrm{eval}}$ is the set of evaluated RL optimizer steps, and $\mathrm{acc}$ is pass@1 on the $3{,}000$ medium-level HotpotQA dev questions used for checkpoint selection (\S\ref{app:HotpotQA-checkpoint-selection}). We compare $c_{\mathrm{RL}}(i^\star)$ with $c_{\mathrm{OP}}$, the cost of producing $\pi_2$ from $\pi_1$.

These runs predate our final SFT pipeline and the dev-set checkpoint selection of \S\ref{app:HotpotQA-checkpoint-selection}. The $\pi_1$ and $\pi_2$ used here are earlier checkpoints of the same two recipes, both produced by that earlier pipeline, so their dev accuracies are lower than the selected entries of Tables~\ref{app:hotpotqa-pure-bc-train-detail} and~\ref{app:hotpotqa-train-detail}.

\subsubsection*{Cost of \OPshort{}}

Producing $\pi_2$ takes three stages beyond $\pi_1$. First, $\pi_1$ rolls out three times on each of the $1{,}000$ training questions. Second, the teacher API continues each rollout for one turn. Third, the base student $\pi_0$ is trained by SFT on the demonstrations and the continuations. Rollouts and SFT run on the node, and we estimate their durations from trainer logs' timestamps. The teacher is a paid API, so we convert its incremental expense over the demonstrations that $\pi_1$ already used, \$0.67 under $0.1\times$ cache-read pricing, into node time at an assumed cloud rental rate of \$20 per node-hour:
\begin{equation}
t_{\mathrm{API}} = \frac{\$0.67}{\$20 \text{ per node-hour}} \approx 2 \text{ minutes}.
\end{equation}
Table~\ref{tab:ablation-rl-opcost} sums the three stages to $c_{\mathrm{OP}} \approx 0.37$ node-hours.

\begin{table}[htbp]
\centering
\small
\caption{Cost of producing the \OPshort{} ($\Ki=1$) student from $\pi_1$ on HotpotQA, in minutes of one $8\times$H100 node.}
\label{tab:ablation-rl-opcost}
\begin{tabular}{llc}
\toprule
Stage & Basis & Node-minutes \\
\midrule
Student rollouts ($3{,}000$) & Logged timestamps & 4.5 \\
Teacher continuations        & \$0.67 of API calls, at \$20 / 8xH100 node-hour     & 2 \\
SFT                          & Logged timestamps & 15 \\
\midrule
Total                        &                                               & $\approx 22$ ($0.37$ node-hours) \\
\bottomrule
\end{tabular}
\end{table}

\subsubsection*{RL runs}

We run two RL variants from $\pi_1$, GRPO \citep{deepseekmath} and Dr.GRPO \citep{drgrpo}, once each on a single $8\times$H100 node. The RL training set is a pool of $4{,}096$ medium-level HotpotQA questions, disjoint from the $1{,}000$ questions used for SFT and from the dev and test questions, so RL trains on questions that \OPshort{} never sees. Each RL optimizer step takes a batch of 256 questions from this pool, samples 8 rollouts per question, and scores every rollout with an exact-match reward. The two runs differ only in loss aggregation and KL regularization, and Table~\ref{tab:hotpotqa-rl-hparams} lists the full configuration. The search tool, system prompt, and tool-call format are those of SFT data collection (\S\ref{app:HotpotQA-data-setup}). Both runs are evaluated on the dev questions before training and every 4 RL optimizer steps, with one rollout per question at temperature 1.

The trainer logs the duration of every phase of each optimizer step. $c_{\mathrm{RL}}$ sums the training phases, which are rollout generation, reward and advantage computation, log-probability computation, and the policy update. It excludes checkpointing and validation, which takes about a quarter of the wall-clock time in both RL runs.

\begin{table}[htbp]
\centering
\caption{GRPO and Dr.GRPO training configuration on HotpotQA. A single value applies to both runs.}
\label{tab:hotpotqa-rl-hparams}
\begin{tabular}{lp{0.6\linewidth}}
\toprule
\textbf{Hyperparameter} & \textbf{Value} \\
\midrule
Reward                        & 1 if the final answer exactly matches a gold answer after normalization, else 0 \\
Advantage                     & Reward minus the mean reward of the 8 rollouts of the same question, without standard-deviation normalization \\
Policy loss                   & PPO-clip surrogate, $\epsilon = 0.2$ \\
Gradient updates per batch    & 1 \\
Loss aggregation              & \textbf{GRPO}: token mean. \textbf{Dr.GRPO}: per-sequence token sum divided by a fixed length of $3{,}000$ tokens \\
KL regularization             & \textbf{GRPO}: KL loss to the initial $\piOne$, coefficient $\beta = 0.001$. \textbf{Dr.GRPO}: none \\
\midrule
Optimizer                     & AdamW \\
Learning rate                 & $1 \times 10^{-6}$ \\
LR scheduler                  & Constant, no warmup \\
Optimizer betas               & $(0.9, 0.999)$ \\
Weight decay                  & 0.01 \\
Gradient clipping (max norm)  & 1.0 \\
RL training questions         & $4{,}096$ (medium level) \\
Train batch size              & 256 questions $\times$ 8 rollouts $= 2{,}048$ trajectories \\
Micro batch size / GPU        & 8 \\
Precision                     & BF16 \\
\midrule
Student sampling              & Temperature 1.0, top-$p$ 1.0 \\
Max assistant turns per episode & 5 \\
Max first-prompt length       & $2{,}048$ tokens \\
Max episode tokens after the first prompt & $3{,}000$ \\
\bottomrule
\end{tabular}
\end{table}

\subsubsection*{Results}

\begin{figure}[htbp]
\centering
\includegraphics[width=0.72\linewidth]{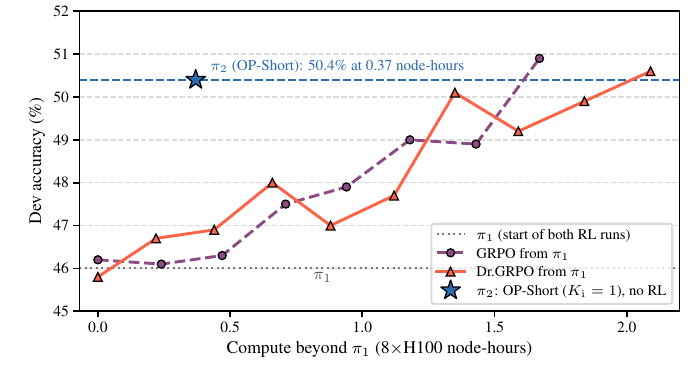}
\caption{HotpotQA dev accuracy of GRPO and Dr.GRPO from $\pi_1$ against RL training compute. The star marks $\pi_2$, the \OPshort{} ($\Ki=1$) student, which reaches $50.4\%$ without RL at $0.37$ node-hours beyond $\pi_1$. The dotted line is the mean of the two runs' evaluations of $\pi_1$ before any RL update.}
\label{fig:ablation-rl}
\end{figure}

Figure~\ref{fig:ablation-rl} plots dev accuracy against $c_{\mathrm{RL}}$, and Table~\ref{tab:ablation-rl-steps} lists the underlying values. The first evaluated checkpoints above the $50.4\%$ of $\pi_2$ arrive at $c_{\mathrm{RL}}(i^\star) = 1.67$ node-hours for GRPO and $2.09$ for Dr.GRPO, about $4.5\times$ and $5.6\times$ $c_{\mathrm{OP}}$. At matched compute, RL is far from this accuracy. RL optimizer step 8 is the first evaluation after RL has spent at least $c_{\mathrm{OP}}$ node-hours. By then, the dev set accuracy of each RL run still has not risen one percentage point above $\pi_1$'s.

\begin{table}[htbp]
\centering
\small
\caption{RL from $\pi_1$ on HotpotQA: cumulative training node-hours and dev accuracy (\%) at each evaluated RL optimizer step. Bold marks the first checkpoint above the $50.4\%$ of \OPshort{}.}
\label{tab:ablation-rl-steps}
\setlength{\tabcolsep}{4pt}
\begin{tabular}{ll*{10}{c}}
\toprule
& RL optimizer step & 0 & 4 & 8 & 12 & 16 & 20 & 24 & 28 & 32 & 36 \\
\midrule
\multirow{2}{*}{GRPO}
& node-hours & 0 & 0.24 & 0.47 & 0.71 & 0.94 & 1.18 & 1.43 & \textbf{1.67} & & \\
& accuracy   & 46.2 & 46.1 & 46.3 & 47.5 & 47.9 & 49.0 & 48.9 & \textbf{50.9} & & \\
\midrule
\multirow{2}{*}{Dr.GRPO}
& node-hours & 0 & 0.22 & 0.44 & 0.66 & 0.88 & 1.12 & 1.35 & 1.59 & 1.84 & \textbf{2.09} \\
& accuracy   & 45.8 & 46.7 & 46.9 & 48.0 & 47.0 & 47.7 & 50.1 & 49.2 & 49.9 & \textbf{50.6} \\
\bottomrule
\end{tabular}
\end{table}

\clearpage
\section{Example Rollout: HotpotQA}
\label{app:rollout-anatomy}

This appendix illustrates how a ReAct-style rollout decomposes into
actions, observations, and tokens, instantiating the formalism in
\S\ref{sec:methods}.
Figure~\ref{fig:rollout-detail} shows a single rollout on one of our
HotpotQA training questions: a system prompt and question followed by
alternating agent turns (thinking and tool-call segments) and
environment observations (tool responses), with token IDs annotated
so the next-token-prediction target is explicit.
Figure~\ref{fig:react-schematic} re-draws the same trajectory as a
linearized sequential decision process: the policy $\pi_\theta$ emits
each action $a_t$ autoregressively conditioned on the growing history
$h_{t-1} = (q, a_1, o_1, \ldots, a_{t-1}, o_{t-1})$, and the dashed
box marks the next-step prediction target supervised at training
time.\newline

\begin{figure}[ht]
\centering
\resizebox{0.95\linewidth}{!}{%
\begin{tikzpicture}[
    seg/.style={minimum height=14pt, inner xsep=3pt, inner ysep=2pt,
                font=\small, anchor=north west},
    tseg/.style={seg, font=\scriptsize\ttfamily},
    qfill/.style={fill=qGray},
    tfill/.style={fill=thinkLav},
    afill/.style={fill=actionLav},
    efill/.style={fill=envGreen},
    splbl/.style={font=\normalsize, anchor=east, text=black!70},
    timearrow/.style={-{Stealth[length=6pt]}, timeBlue, line width=1.6pt},
]
\def\tokgap{2pt}
\def\spangap{10pt}

\node[seg, qfill] (Qt) at (0,0)
    {[Sys: You are a helpful assistant \dots\ search tool \dots]\
    Where was the draft that Chad Johnson was drafted held at?};
\node[tseg, qfill] (Qk) at ([yshift=-\tokgap]Qt.south west)
    {2610\ 525 \dots\ 10967\ 572\ 279\ 9960\ 429\ 42030\ 11351\ 572\ 37956\ 5644\ 518\ 30};
\node[splbl] at ($(Qt.north west)!0.5!(Qk.south west)+(-10pt,0)$) {$q$};

\node[seg, tfill] (A1tt) at ([yshift=-\spangap]Qk.south west)
    {I need to find where Chad Johnson was drafted \dots};
\node[seg, afill, font=\footnotesize] (A1ta) at (A1tt.north east)
    {\tc\ \dots\ Chad Johnson NFL draft location \dots\ \tce};
\node[tseg, tfill] (A1kt) at ([yshift=-\tokgap]A1tt.south west)
    {40\ 1184\ 311\ 1477\ 1380\ 42030\ 11351\ 572\ 37956 \dots};
\node[tseg, afill] (A1ka) at (A1kt.north east)
    {151657 \dots\ 1143\ 329\ 11351\ 12588\ 9960\ 3728 \dots\ 151658};
\node[splbl] at ($(A1tt.north west)!0.5!(A1kt.south west)+(-10pt,0)$) {$\boldsymbol{a_1}$};

\node[seg, efill] (O1t) at ([yshift=-\spangap]A1kt.south west)
    {\tr\ \dots\ Doc\,1 \dots\ born in Miami \dots\
    Doc\,2 \dots\ NFL Draft \dots\ \tre};
\node[tseg, efill] (O1k) at ([yshift=-\tokgap]O1t.south west)
    {27\ 14172\ 9655\ 397 \dots\ 9550\ 220\ 16 \dots\
    9223\ 304\ 17584 \dots\ 9550\ 220\ 17 \dots\
    12588\ 28564 \dots\ 522\ 14172\ 9655\ 29};
\node[splbl] at ($(O1t.north west)!0.5!(O1k.south west)+(-10pt,0)$) {$o_1$};

\node[seg, tfill] (A2tt) at ([yshift=-\spangap]O1k.south west)
    {Chad Johnson was drafted \dots\ held at \dots};
\node[seg, afill] (A2ta) at (A2tt.north east)
    {\ans\ Radio City Music Hall, New York City \anse};
\node[tseg, tfill] (A2kt) at ([yshift=-\tokgap]A2tt.south west)
    {1143\ 329\ 11351\ 572\ 37956 \dots\ 5644\ 518 \dots};
\node[tseg, afill] (A2ka) at (A2kt.north east)
    {27\ 9217\ 29\ 13484\ 4311\ 10717\ 10926\ 11\ 1532\ 4261\ 4311\ 690\ 9217\ 29};
\node[splbl] at ($(A2tt.north west)!0.5!(A2kt.south west)+(-10pt,0)$) {$\boldsymbol{a_2}$};

\draw[timearrow]
    ([xshift=-34pt]Qt.north west) -- ([xshift=-34pt]A2kt.south west);
\node[rotate=-90, font=\small, text=timeBlue, anchor=center]
    at ([xshift=-42pt]$(Qt.north west)!0.5!(A2kt.south west)$)
    {Time steps $t=0,1,2,\dots$};
\end{tikzpicture}
}%

\vspace{6pt}
\noindent\small
\colorbox{qGray}{\strut\hspace{6pt}} Task query ($q$) \qquad
\colorbox{thinkLav}{\strut\hspace{6pt}} Agent reasoning \qquad
\colorbox{actionLav}{\strut\hspace{6pt}} Agent action \qquad
\colorbox{envGreen}{\strut\hspace{6pt}} Env.\ response ($o$)

\caption{A real ReAct-style rollout instantiating the sequential decision process in actual model output and token IDs. For each span, the top row shows abbreviated text and the bottom row shows corresponding token IDs (under a Qwen\,2.5 tokenizer). The agent emits a reasoning trace and tool call (together as $a_1$), receives a search result ($o_1$), then boxes a final answer in $a_2$. XML tags such as \tc/\tce{} delimit span boundaries in the raw sequence.}
\label{fig:rollout-detail}
\end{figure}
\vspace{1em}
\begin{figure}[ht]
\centering
\includegraphics[width=\textwidth]{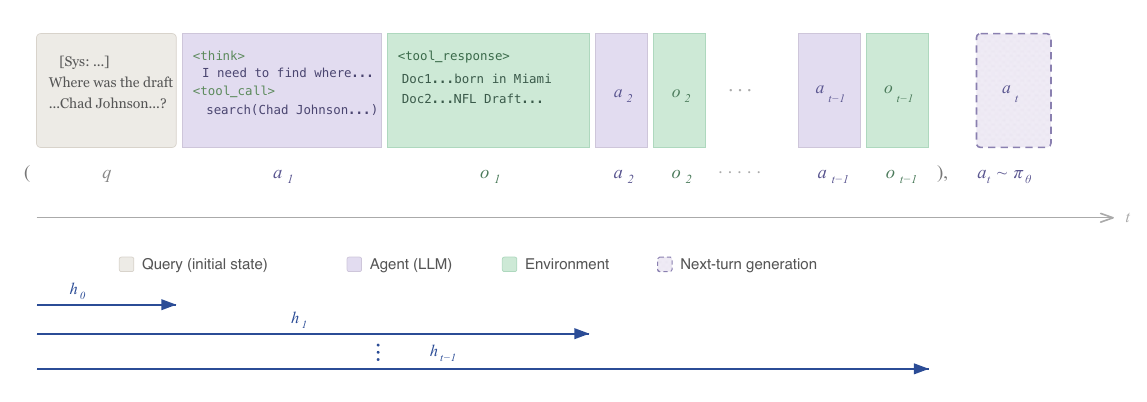}
\caption{Linearized schematic of the Figure \ref{fig:rollout-detail} rollout trajectory as a sequential decision process. The agent's next assistant turn generation, $a_t$, is sampled token by token, autoregressively from the conditional policy distribution $\pi_\theta(\cdot | h_{t-1})$.}
\label{fig:react-schematic}
\end{figure}
\clearpage
\section{Example Trajectory: OpenThoughts-Agent}
\label{app:op-example}

\begingroup\setlength{\emergencystretch}{2em}%
We show one hybrid trajectory $\tau'$ from the on-policy (OP) data $\mathcal{D}_{\mathrm{OP}}$ collected for the OpenThoughts-Agent/TB-Dev experiments. %
The student is \mbox{Qwen3-8B} and the teacher $\pi_e$ is \mbox{GLM-4.6}.
The trajectory prefix $\tau|_{t<t'}$ is generated by the student policy $\pi_1 = \mathrm{SFT}(\pi_0, 10\%\,\mathcal{D}_{\mathrm{BC}})$, which is held fixed during collection.
At switch time $t' = 4$, the teacher branches from the rollout snapshot $x_3(\tau)$, so that $x_3(\tau') = x_3(\tau)$, and continues until it indicates task completion at $T(\tau') = 6$.
This teacher continuation has three turns.
The task $q \in \mathcal{Q}_{\mathrm{train}}$ is an nl2bash task from the OpenThoughts-Agent training pool.

\paragraph{Why this trajectory.}
The task asks for file names \emph{without the leading} \verb!./!.
At $t = 3$, the student's plan restates this requirement, but its action $a_3$ runs \verb!find . -maxdepth 1 -type f! \verb!| sort -r!, which keeps the prefix, and $o_3$ shows \verb!./file2.log!, \verb!./file1.txt!, and \verb!./README! in the capture file.
The teacher's first action $a^\star_4$ reads this violation off $o_3$ and switches to \verb!-printf '%f\n'!; $o'_4$ shows the corrected capture file, and the remaining two teacher turns verify the file and confirm task completion.
This proposal contributes the student context $h_3(\tau)$ paired with the retained teacher continuation $(a^\star_4, o'_4, a^\star_5, o'_5, a^\star_6, o'_6)$ to $\mathcal{D}_{\mathrm{OP}}$; only the teacher-generated actions are SFT targets.
The full teacher demonstration for the same task in $\mathcal{D}_{\mathrm{BC}}$ issues the correct command directly and never visits a context like $h_3(\tau)$, so supervision on recovering from this student error comes only from the OP record, whose context is generated by the student.

\paragraph{Partial observability.}
The environment state $s_t$ is the sandbox (its file system and shell session), whereas the policy samples $a_t \sim \pi(\cdot \mid h_{t-1})$ from the LLM context alone, so any part of $s_t$ not shown in $h_t$ is unobservable for $\pi$.
Here $h_0 = (q)$ contains the instructions, the task description, and an empty terminal screen; it does not list the files in \verb!/workspace! or state whether \verb!/output! exists.
The student's first action $a_1$ (\verb!ls -la!) gathers information: $o_1$ reveals the top level of \verb!/workspace!, while the contents of \verb!subdir/! remain unobserved throughout $\tau'$.
The student never checks whether \verb!/output! exists and instead runs the idempotent \verb!mkdir -p!.
At $t = 3$, $a_3$ writes to a file rather than the terminal, so the incorrect file content is part of $s_3$ and enters $h_3$ only through the harness's side-channel preview (\verb!Sidechannel Output!) in $o_3$.
At the switch, the teacher branches from the rollout snapshot $x_3(\tau) = (h_3(\tau), s_3(\tau))$: it reads the student context $h_3(\tau)$ and acts in the same sandbox, where $a^\star_4$ overwrites the incorrect capture file the student left in $s_3$.

\paragraph{How to read the figure.}
The strip below summarizes the timeline of $\tau'$, and box colors follow the roles in $\tau'$.
Lines shaded \tbswatch{tbErrHL}{red} mark the student error, lines shaded \tbswatch{tbFixHL}{green} mark the teacher's correction, and lines shaded \tbswatch{tbReqHL}{yellow} state the requirement the student violates.
The black banner marks the switch time $t'$.
Each action box names the policy that samples the action; throughout, $h_t$ denotes $h_t(\tau')$, which equals $h_t(\tau)$ for $t \le 3$.
\par\endgroup

\begin{tbtimeline}
\tbchip{tbPrompt}{$q$}\,\tbarrow\,%
\tbchip{tbStudent}{$a_1$}\tbchip{tbEnv}{$o_1$}\,\tbarrow\,%
\tbchip{tbStudent}{$a_2$}\tbchip{tbEnv}{$o_2$}\,\tbarrow\,%
\tbchip{tbStudent}{$a_3$\,\tbcrossW}\tbchip{tbEnv}{$o_3$}\,\tbarrow\,%
\tbswitchchip{$t'{=}4$}%
\tbchip{tbTeacher}{$a^\star_4$\,\tbcheckW}\tbchip{tbEnv}{$o'_4$}\,\tbarrow\,%
\tbchip{tbTeacher}{$a^\star_5$}\tbchip{tbEnv}{$o'_5$}\,\tbarrow\,%
\tbchip{tbTeacher}{$a^\star_6$}\tbchip{tbEnv}{$o'_6$}\par\vspace{12pt}
{\footnotesize
\tblegend{tbPrompt}{task $q$}\qquad
\tblegend{tbStudent}{student action $a_t$ ($\pi_1$)}\qquad
\tblegend{tbEnv}{observation $o_t$, $o'_t$}\qquad
\tblegend{tbTeacher}{teacher action $a^\star_t$ ($\pi_e$)}\par}
\end{tbtimeline}

\begin{taskbox}[tbprompt]{Task $q$ (initial context $h_0 = (q)$)}
\begin{trajverb}[highlightlines={56}, highlightcolor=tbReqHL]
You are an AI assistant tasked with solving command-line tasks in a Linux environment. You will be given a task description and the output from previously executed commands. Your goal is to solve the task by providing batches of shell commands.

Format your response as JSON with the following structure:

{
  "analysis": "Analyze the current state based on the terminal output provided. What do you see? What has been accomplished? What still needs to be done?",
  "plan": "Describe your plan for the next steps. What commands will you run and why? Be specific about what you expect each command to accomplish.",
  "commands": [
    {
      "keystrokes": "ls -la\n",
      "duration": 0.1
    },
    {
      "keystrokes": "cd project\n",
      "duration": 0.1
    }
  ],
  "task_complete": true
}

Required fields:
- "analysis": Your analysis of the current situation
- "plan": Your plan for the next steps
- "commands": Array of command objects to execute

Optional fields:
- "task_complete": Boolean indicating if the task is complete (defaults to false if not present)

Command object structure:
- "keystrokes": String containing the exact keystrokes to send to the terminal (required)
- "duration": Number of seconds to wait for the command to complete before the next command will be executed (defaults to 1.0 if not present)

IMPORTANT: The text inside "keystrokes" will be used completely verbatim as keystrokes. Write commands exactly as you want them sent to the terminal:
- You must end every command with a newline (\n) or it will not execute.
- For special key sequences, use tmux-style escape sequences:
  - C-c for Ctrl+C
  - C-d for Ctrl+D

The "duration" attribute specifies the number of seconds to wait for the command to complete (default: 1.0) before the next command will be executed. On immediate tasks (e.g., cd, ls, echo, cat) set a duration of 0.1 seconds. On commands (e.g., gcc, find, rustc) set a duration of 1.0 seconds. On slow commands (e.g., make, python3 [long running script], wget [file]) set an appropriate duration as you determine necessary.

It is better to set a smaller duration than a longer duration. It is always possible to wait again if the prior output has not finished, by running {"keystrokes": "", "duration": 10.0} on subsequent requests to wait longer. Never wait longer than 60 seconds; prefer to poll to see intermediate result status.

Important notes:
- Each command's keystrokes are sent exactly as written to the terminal
- Do not include extra whitespace before or after the keystrokes unless it's part of the intended command
- Extra text before or after the JSON will generate warnings but be tolerated
- The JSON must be valid - use proper escaping for quotes and special characters within strings
- Commands array can be empty if you want to wait without taking action

Task Description:
# Terminal Automation Request

**TERMINAL ENVIRONMENT**: Ubuntu 24.04 with bash, git, curl, wget, Python 3.10, pip, coreutils, and build-essential.

## Goal
List regular files only in the current directory (do not descend into subdirectories), output their names without leading ./, sort the list in reverse order, and save both stdout and stderr produced by the command into the capture file.

## Requirements
- From `/workspace`, assemble any missing fixtures using standard shell utilities before running your command.
- Execute the command(s) necessary to satisfy the goal in a single shell session so that both stdout and stderr are captured.
- The combined output must be written to `/output/command_capture.txt`.
- Do not delete any helper files generated during execution; the verifier inspects them.

After completing the workflow, confirm that `/output/command_capture.txt` exists and contains the captured output. Return a short confirmation summarising the exit status and file size.

## Available Tools
- bash / sh / coreutils
- git, curl, wget
- python3.10 and pip
- build-essential (gcc, make)

<!-- OT_ROLLOUT_TAG: rollout=003 task_id=task_1246__rollout003 -->

Current terminal state:
Current Terminal Screen:
root@b1fdf461-e987-4e4f-9f15-7b863bd17f95:/workspace#
\end{trajverb}
\end{taskbox}

\begin{taskbox}[tbstudent]{$t = 1$ \textbullet\ Student action $a_{1} \sim \pi_1(\cdot \mid h_{0})$}
\begin{trajverb}
{
  "analysis": "I'm in the /workspace directory and need to list regular files only (no directories), sort them in reverse order, and capture both stdout and stderr to /output/command_capture.txt. I should first check what's in the current directory and then ensure the /output directory exists before running the final command.",
  "plan": "1. First, check the current directory contents to understand what files exist\n2. Ensure the /output directory exists\n3. Execute a command that lists regular files only, sorts them in reverse order, and captures both stdout and stderr to the required output file\n4. Verify the capture file was created and contains the expected output I will execute one command first, inspect the result, and continue step by step.",
  "commands": [
    {
      "keystrokes": "ls -la\n",
      "duration": 1.0
    }
  ],
  "task_complete": false
}
\end{trajverb}
\end{taskbox}

\begin{taskbox}[tbenv]{Observation $o_{1}$}
\begin{trajverb}
New Terminal Output:
root@b1fdf461-e987-4e4f-9f15-7b863bd17f95:/workspace# ls -la
total 0
drwxr-xr-x 1 root root  88 Apr 25 19:02 .
drwxr-xr-x 1 root root 102 Apr 26 01:20 ..
-rw-r--r-- 1 root root   0 Jan  1  1970 README
-rw-r--r-- 1 root root   0 Jan  1  1970 file1.txt
-rw-r--r-- 1 root root   0 Jan  1  1970 file2.log
drwxr-xr-x 2 root root  31 Apr 25 19:02 subdir
root@b1fdf461-e987-4e4f-9f15-7b863bd17f95:/workspace#
\end{trajverb}
\end{taskbox}

\begin{taskbox}[tbstudent]{$t = 2$ \textbullet\ Student action $a_{2} \sim \pi_1(\cdot \mid h_{1})$}
\begin{trajverb}
{
  "analysis": "I can see the current directory contains 3 regular files (README, file1.txt, file2.log) and a subdirectory called 'subdir'. I need to list only the regular files, sort them in reverse order, and capture both stdout and stderr to /output/command_capture.txt. I also need to ensure the /output directory exists.",
  "plan": "1. Create the /output directory if it doesn't exist\n2. Execute a command that finds regular files only in the current directory, sorts them in reverse order, and captures both stdout and stderr to /output/command_capture.txt\n3. Verify the capture file was created and contains the expected output I will execute one command first, inspect the result, and continue step by step.",
  "commands": [
    {
      "keystrokes": "mkdir -p /output\n",
      "duration": 1.0
    }
  ],
  "task_complete": false
}
\end{trajverb}
\end{taskbox}

\begin{taskbox}[tbenv]{Observation $o_{2}$}
\begin{trajverb}
New Terminal Output:

root@b1fdf461-e987-4e4f-9f15-7b863bd17f95:/workspace# mkdir -p /output
root@b1fdf461-e987-4e4f-9f15-7b863bd17f95:/workspace#
\end{trajverb}
\end{taskbox}

\begin{taskbox}[tbstudent]{$t = 3$ \textbullet\ Student action $a_{3} \sim \pi_1(\cdot \mid h_{2})$\quad\tbtagerr}
\renewcommand{\tbMixColor}{\ifnum\value{FancyVerbLine}=3 tbReqHL\else tbErrHL\fi}
\begin{trajverb}[highlightlines={3,6}, highlightcolor=\tbMixColor]
{
  "analysis": "The /output directory has been created successfully. Now I need to execute the main command that lists regular files only in the current directory (not descending into subdirectories), sorts them in reverse order, and captures both stdout and stderr to /output/command_capture.txt.",
  "plan": "Execute a command that uses find to locate regular files only in the current directory (maxdepth 1), sorts them in reverse order, and redirects both stdout and stderr to /output/command_capture.txt. The command should capture the file names without leading './'.",
  "commands": [
    {
      "keystrokes": "find . -maxdepth 1 -type f | sort -r > /output/command_capture.txt 2>&1\n",
      "duration": 1.0
    }
  ],
  "task_complete": false
}
\end{trajverb}
\end{taskbox}

\begin{taskbox}[tbenv]{Observation $o_{3}$}
\begin{trajverb}[highlightlines={10,11,12}, highlightcolor=tbErrHL]
New Terminal Output:

root@b1fdf461-e987-4e4f-9f15-7b863bd17f95:/workspace# find . -maxdepth 1 -type f | sort -r > /output/command_capture.txt 2>&1
root@b1fdf461-e987-4e4f-9f15-7b863bd17f95:/workspace#

Sidechannel Output:
[sidechannel] /output/command_capture.txt exists
-rw-r--r-- 1 root root 33 Apr 26 01:21 /output/command_capture.txt
[sidechannel] preview
./file2.log
./file1.txt
./README
\end{trajverb}
\end{taskbox}

\begin{tbcallout}[tberrcallout]
\tbcross\ \textbf{Student error.} The task requires names without the leading \verb!./!, and the plan of $a_3$ restates this (yellow), but \verb!find . -maxdepth 1 -type f! prints paths relative to \verb!.!; the capture file in $o_3$ therefore contains \verb!./file2.log!, \verb!./file1.txt!, \verb!./README! (red lines).
\end{tbcallout}

\begin{tbswitch}
\tbswitchicon\ \textbf{SWITCH at $t' = 4$: the teacher $\pi_e$ (GLM-4.6) branches from the rollout snapshot $x_3(\tau)$.}\\[1pt]
{\small Everything below is the teacher continuation; only the teacher actions $a^\star_4, a^\star_5, a^\star_6$ are SFT targets.}
\end{tbswitch}

\begin{taskbox}[tbteacher]{$t = 4$ \textbullet\ Teacher action $a^\star_{4} \sim \pi_e(\cdot \mid h_{3})$\quad\tbtagfix}
\begin{trajverb}[highlightlines={2,6}, highlightcolor=tbFixHL]
{
  "analysis": "The command executed successfully and created the capture file. However, I notice the output contains './' prefixes which the requirements specified should not be included. I need to modify the command to remove the './' prefix from the file names.",
  "plan": "Execute a corrected command that removes the './' prefix from the file names while still listing only regular files, sorting in reverse order, and capturing both stdout and stderr to the output file.",
  "commands": [
    {
      "keystrokes": "find . -maxdepth 1 -type f -printf '%f\\n' | sort -r > /output/command_capture.txt 2>&1\n",
      "duration": 1.0
    }
  ],
  "task_complete": false
}
\end{trajverb}
\end{taskbox}

\begin{taskbox}[tbenv]{Observation $o'_{4}$}
\begin{trajverb}[highlightlines={10,11,12}, highlightcolor=tbFixHL]
New Terminal Output:

root@b1fdf461-e987-4e4f-9f15-7b863bd17f95:/workspace# find . -maxdepth 1 -type f -printf '%f\n' | sort -r > /output/command_capture.txt 2>&1
root@b1fdf461-e987-4e4f-9f15-7b863bd17f95:/workspace#

Sidechannel Output:
[sidechannel] /output/command_capture.txt exists
-rw-r--r-- 1 root root 27 Apr 26 01:21 /output/command_capture.txt
[sidechannel] preview
file2.log
file1.txt
README
\end{trajverb}
\end{taskbox}

\begin{tbcallout}[tbfixcallout]
\tbcheck\ \textbf{Teacher correction.} $a^\star_4$ detects the violation from $o_3$ and rewrites the command with \verb!-printf '%f\n'!, which prints bare file names; $o'_4$ shows the capture file \verb!file2.log!, \verb!file1.txt!, \verb!README! (green lines). $a^\star_5$ and $a^\star_6$ verify the file and confirm task completion.
\end{tbcallout}

\begin{taskbox}[tbteacher]{$t = 5$ \textbullet\ Teacher action $a^\star_{5} \sim \pi_e(\cdot \mid h_{4})$}
\begin{trajverb}
{
  "analysis": "Perfect! The command executed successfully and the capture file now contains the regular files from the current directory (file2.log, file1.txt, README) sorted in reverse order without the './' prefix. The file size is 27 bytes and exists at /output/command_capture.txt. Both stdout and stderr were captured as required.",
  "plan": "Verify the final result by checking the capture file exists and contains the expected output, then confirm task completion. I will execute one command first, inspect the result, and continue step by step.",
  "commands": [
    {
      "keystrokes": "ls -la /output/command_capture.txt\n",
      "duration": 1.0
    }
  ],
  "task_complete": true
}
\end{trajverb}
\end{taskbox}

\begin{taskbox}[tbenv]{Observation $o'_{5}$}
\begin{trajverb}
Current terminal state:
New Terminal Output:

root@b1fdf461-e987-4e4f-9f15-7b863bd17f95:/workspace# ls -la /output/command_capture.txt
-rw-r--r-- 1 root root 27 Apr 26 01:21 /output/command_capture.txt
root@b1fdf461-e987-4e4f-9f15-7b863bd17f95:/workspace#

Sidechannel Output:
[sidechannel] /output/command_capture.txt exists
-rw-r--r-- 1 root root 27 Apr 26 01:21 /output/command_capture.txt
[sidechannel] preview
file2.log
file1.txt
README

Are you sure you want to mark the task as complete? This will trigger your solution to be graded and you won't be able to make any further corrections. If so, include "task_complete": true in your JSON response again.
\end{trajverb}
\end{taskbox}

\begin{taskbox}[tbteacher]{$t = 6$ \textbullet\ Teacher action $a^\star_{6} \sim \pi_e(\cdot \mid h_{5})$\quad\tbtagend}
\begin{trajverb}
{
  "analysis": "The task has been completed successfully. The capture file /output/command_capture.txt exists with 27 bytes and contains the regular files (file2.log, file1.txt, README) sorted in reverse order without './' prefixes. Both stdout and stderr were captured as required. The file verification confirms the task requirements have been met.",
  "plan": "Final confirmation - the task is complete with the capture file containing the correctly formatted output.",
  "commands": [],
  "task_complete": true
}
\end{trajverb}
\end{taskbox}

\begin{taskbox}[tbenv]{Observation $o'_{6}$}
\begin{trajverb}
New Terminal Output:

root@b1fdf461-e987-4e4f-9f15-7b863bd17f95:/workspace#

Sidechannel Output:
[sidechannel] /output/command_capture.txt exists
-rw-r--r-- 1 root root 27 Apr 26 01:21 /output/command_capture.txt
[sidechannel] preview
file2.log
file1.txt
README
\end{trajverb}
\end{taskbox}

\tbfigcaption{Hybrid trajectory $\tau'$ for an nl2bash task $q \in \mathcal{Q}_{\mathrm{train}}$ in the OpenThoughts-Agent OP data $\mathcal{D}_{\mathrm{OP}}$. Purple: task $q$. Orange: student actions $a_1$--$a_3$ from the student policy $\pi_1$. Blue: observations. Green: teacher actions $a^\star_4$--$a^\star_6$ from the teacher $\pi_e$ (GLM-4.6) after the switch at $t' = 4$. Red-shaded lines mark the student error, green-shaded lines the teacher's correction, and yellow-shaded lines the requirement the student violates.}{fig:op-example}

\clearpage

\end{document}